\documentclass[a4paper,12pt]{article}

\usepackage{comment}
\usepackage{fancyvrb}
\usepackage{graphicx}
\usepackage{hyperref}
\usepackage[ragged]{footmisc}
\usepackage{ragged2e}
\usepackage{everysel}
\usepackage[section]{placeins}
\usepackage{layout}
\usepackage{floatpag}
\usepackage{gensymb}
\usepackage{lscape}
\usepackage{geometry}
\usepackage{lscape}

\begin{document}

\bibliographystyle{plain}

\title{Commonsense Reasoning, Commonsense Knowledge, and The SP Theory of Intelligence}

\author{J Gerard Wolff\footnote{Dr Gerry Wolff, BA (Cantab), PhD (Wales), CEng, MBCS, MIEEE; CognitionResearch.org, Menai Bridge, UK; \href{mailto:jgw@cognitionresearch.org}{jgw@cognitionresearch.org}; +44 (0) 1248 712962; +44 (0) 7746 290775; {\em Skype}: gerry.wolff; {\em Web}: \href{http://www.cognitionresearch.org}{www.cognitionresearch.org}.}}

\maketitle

\begin{abstract}

This paper describes how the {\em SP Theory of Intelligence} with the {\em SP Computer Model}, outlined in an Appendix, may throw light on aspects of commonsense reasoning (CSR) and commonsense knowledge (CSK), as discussed in another paper by Ernest Davis and Gary Marcus (DM). In four main sections, the paper describes: 1) The main problems to be solved; 2) Other research on CSR and CSK; 3) Why the SP system may prove useful with CSR and CSK 4) How examples described by DM may be modelled in the SP system. With regard to successes in the automation of CSR described by DM, the SP system's strengths in simplification and integration may promote seamless integration across these areas, and seamless integration of those area with other aspects of intelligence. In considering challenges in the automation of CSR described by DM, the paper describes in detail, with examples of SP-multiple-alignments. how the SP system may model processes of interpretation and reasoning arising from the horse's head scene in {\em The Godfather} film. A solution is presented to the `long tail' problem described by DM. The SP system has some potentially useful things to say about several of DM's objectives for research in CSR and CSK.

\end{abstract}


\section{Introduction}\label{introduction_section}

``AI has seen great advances of many kinds recently, but there is one critical area where progress has been extremely slow: ordinary commonsense.'' So say Ernest Davis and Gary Marcus \cite[p.~92]{davis_marcus_2015}, illustrating the point with questions that are easy for people to answer but can be difficult for artificial systems, such as ``Who is taller, Prince William or his baby son Prince George'' and ``Can you make a salad out of a polyester shirt?'' \cite[pp.~92--93]{davis_marcus_2015}.

As further illustration, they write: ``If you read the text, `I stuck a pin in a carrot; when I pulled the pin out, it had a hole,' you need not consider the possibility `it' refers to the pin.'' and ``Anyone who has seen the unforgettable horse's head scene in {\em The Godfather} immediately realizes what is going on. It is not just it is unusual to see a severed horse head, it is clear Tom Hagen is sending Jack Woltz a message---if I can decapitate your horse, I can decapitate you; cooperate, or else. For now, such inferences lie far beyond anything in artificial intelligence.'' \cite[p.~93]{davis_marcus_2015}.

Issues like those can be challenging for AI but the {\em SP theory of Intelligence} and its realisation in the {\em SP Computer Model}, outlined in Appendix \ref{outline_of_sp_system_appendix}, may provide a way forward. This paper describes how the SP Theory and Computer Model (which, together, will be referred to as the ``SP system'') may prove useful in modelling commonsense reasoning and commonsense knowledge as described by Davis and Marcus.

For the sake of brevity in this paper, commonsense reasoning will be abbreviated as `CSR', commonsense knowledge will be abbreviated as `CSK', and the two together will be abbreviated as `CSRK'. Also, Davis and Marcus, as they write in \cite{davis_marcus_2015}, will be referred to by the first letters of those two names, `DM'.

{\em Readers who are not already familiar with the SP system are urged to read the outline in Appendix \ref{outline_of_sp_system_appendix} before reading the rest of the paper}.

In what follows, there are four main sections:

\begin{itemize}

    \item A very short Section \ref{main_problems_to_be_solved_section}, outlining the main problems to be solved in a comprehensive account of human-level performance in CSRK.

    \item A somewhat longer Section \ref{research_on_cskr_section}, summarising other research on CSRK.

    \item At more length, Section \ref{sp_potential_in_csrk_section} describes why the SP system may provide a foundation for CSRK.

    \item And at still more length, Section \ref{aspects_of_csrk_section} discusses aspects of CSRK and how they may be modelled in the SP system, drawing mainly on examples and discussion in \cite{davis_marcus_2015}.

\end{itemize}

\subsection{Motivation}

Before embarking on the main body of the paper, it is perhaps worth saying a little about the potential benefits of research on commonsense reasoning and commonsense knowledge.

As part of AI, research on CSRK may help to shift attention towards aspects of AI which have been somewhat overshadowed by success in relatively narrow areas such as the playing of games, welcome though such successes undoubtedly are. That shift in attention may also be a helpful reminder that research in AI is hard and that there are still difficult problems to be solved.

In practical terms, progress with CSRK has potential to yield many benefits. For example:

\begin{itemize}

    \item An autonomous robots with human levels of commonsense would probably be able to take over many humdrum chores, but without commonsense, it would be much less useful.

    \item Although commonsense is not normally associated with legal reasoning, an understanding of everyday situations and how they work is an essential part of most legal arguments. Artificial CSRK at human levels has considerable potential to assist in the drafting of legislation and in the preparation of legal cases.

    \item In a similar way, success in the understanding of human-level CSRK has considerable potential to assist with crime prevention and detection.

    \item Success with CSRK has potential to assist in the preparation of plans of many kinds including those in business, in military operations, in administration, and so on.

    \item And DM say ``...~many intelligent tasks, such as understanding texts, computer vision, planning, and scientific reasoning require [commonsense] kinds of real world knowledge and reasoning abilities.'' \cite[p.~92]{davis_marcus_2015}.

\end{itemize}

\section{The Main Problems to be Solved}\label{main_problems_to_be_solved_section}

In broad terms, the achievement of human levels of performance in CSRK requires an understanding of: 1) {\em Reasoning}: CSR and related processes such as pattern recognition. 2) {\em Representation of knowledge}: CSK and how it should be structured to support CSR, and 3) {\em Learning}: how CSK may be learned.

In research on CSRK, these three problems do not have equal status. Reasoning is what delivers the end result, and it may include a specification of the form or forms that CSK should take. But although learning is a very important topic in AI, it is somewhat tangential to the delivery of human-level CSRK, except that it must produce the forms of CSK that are required by CSR.

Because of this asymmetry, it is not surprising to see that the main focus of \cite{davis_marcus_2015} is on CSR: ``In this article, we argue that commonsense
reasoning is important in many AI tasks, from text understanding to computer vision, planning and reasoning, and discuss four specific problems
where substantial progress has been made. We consider why the problem in its general form is so difficult and why progress has been so slow, and
survey various techniques that have been attempted.'' \cite[p.~93]{davis_marcus_2015}. CSK and how it should be structured does feature in the title and body of the paper, but there is less emphasis how CSK may be learned.

As in \cite{davis_marcus_2015}, the main focus here is on CSR, with CSK considered mainly in terms of how it may support CSR.

\section{Other Research on Commonsense Reasoning and Commonsense Knowledge}\label{research_on_cskr_section}

To provide a context for the rest of the paper, this section outlines a selection of research on commonsense reasoning and commonsense knowledge other than what is described in this paper. Here is a small sample of relatively recent studies:

\begin{itemize}

    \item The Cyc project, initiated and led for many years by Douglas Lenat (See, for example, \cite{witbrock_etal_2015,read_lenat_2002}) has assembled a very large database of knowledge about basic concepts and `rules of thumb' about how the world works. The intention has been to facilitate the creation of AI applications that may perform human-like reasoning and which can cope with novel situations that were not preconceived.

        DM write quite extensively about this project \cite[pp.~99--103]{davis_marcus_2015}, with remarks such as ``No systematic evaluation of
        the contents, capacities, and limitations of CYC has been published.'' \cite[p.~101]{davis_marcus_2015}, and ``The [CSRK] field might well benefit if CYC were systematically described and evaluated. If CYC has solved some significant fraction of commonsense reasoning, then it is critical to know that, both as a useful tool, and as a starting point for further research. If CYC has run into difficulties, it would be useful to learn from the mistakes that were made. If CYC is entirely useless, then researchers can at least stop worrying about whether they are reinventing the wheel.'' \cite[p.~103]{davis_marcus_2015}.

    \item `ConceptNet' is a knowledge graph that connects words and phrases of natural language with labeled edges and is designed to represent the general knowledge involved in understanding language (see, for example, \cite{speer_etal_2017}).

    \item `SenticNet' is a three-level knowledge representation for sentiment analysis. The project uses recurrent neural networks to infer primitives by lexical substitution and for grounding common and commonsense knowledge by means of multi-dimensional scaling (see, for example, \cite{cambria_etal_2018}).

    \item Yukun Ma and colleagues (see, for example, \cite{ma_etal_2018}) propose an extension of ``long short-term memory'' (LSTM), termed ``Sentic LSTM''. This augments a LSTM network with an hierarchical  attention  mechanism  comprising  ``target-level attention'' and ``sentence-level attention''. In the system, commonsense knowledge of sentiment-related concepts is incorporated into the  end-to-end  training  of  a  deep  neural  network  for  sentiment classification. Experiments show that the system can outperform state-of-the-art methods in targeted aspect sentiment tasks.

    \item Joseph Blass and Kenneth Forbus \cite{blass_forbus_2017} describe an approach called {\em analogical chaining} to create cognitive systems that can perform commonsense reasoning. In this approach, `commonsense units' are provided to the system via natural language instruction.

    \item Leora Morgenstern and colleagues \cite{morgenstern_etal_2017} discuss plans to run a competition modelled on the `Winograd Schema Challenge', a type of challenge in the interpretation of natural language, described by Terry Winograd \cite{winograd_1972}, which is easy for people but, normally, hard for computers.

    \item Shiqi Zhang and Peter Stone \cite{zhang_stone_2015} discuss aspects of reasoning in intelligent robots, including challenges in modelling the kinds of commonsense reasoning that people find easy. Following a discussion of some alternative frameworks, they introduce the `CORPP' algorithm, with apparent advantages over those alternatives.

    \item Somak Aditya and colleagues \cite{aditya_etal_2015} explore the use of visual commonsense knowledge and other kinds of knowledge for scene understanding. They combine visual processing with techniques from natural language understanding.

    \item Nicole Maslan and colleagues \cite{maslan_etal_2015} present a set of challenge problems for the logical formalization of commonsense knowledge which, unlike other such sets, is designed to support the development of logic-based commonsense theories.

    \item Andr{\'e} Freitas and colleagues \cite{freitas_etal_2014} describe a selective graph-navigation mechanism based on a distributional-relational semantic model which can be applied to querying and reasoning with a variety of knowledge bases, and they discuss how it may be applied.

    \item A paper by Gary Marcus and Ernest Davis \cite{marcus_davis_2013}, a little earlier than \cite{davis_marcus_2015}, discusses an issue related to commonsense reasoning: whether or not the mind should be viewed as a near-optimal or rational engine of probabilistic inference. They argue that this view is markedly less promising than is widely believed. They also argue that the commonly-supposed equivalence between probabilistic inference, on the one hand, and rational or optimal inference, on the other, is not justified.

\end{itemize}

In addition, much other research related to CSRK, published a little earlier than the studies summarised above, is described in \cite{harmelin_etal_2008}. And these sources are by no means exhaustive.

Given the cornucopia of CSRK-related research outlined above, readers may quite reasonably ask: are there not at least one or two that can meet some or all of the challenges described by DM? In answer to that question, DM do indeed describe four approaches that are relatively successful, and those four approaches, with DM's observations, are discussed in this paper in Section \ref{successes_in_csr_section}, below.

\section{Why the SP System May Prove Useful With Commonsense Reasoning and Commonsense Knowledge}\label{sp_potential_in_csrk_section}

Readers may wish to know what justification there is, if any, for the suggestion in the Introduction that the SP system may prove useful with CSRK. This section tries to provide some answers.

\subsection{Simplicity and Power}

The SP system, as noted in Appendix \ref{distinctive_features_and_advantages_appendix}, is the product of a unique attempt to simplify and integrate observations and concepts across a wide area---to develop a system that combines conceptual {\em Simplicity} with descriptive or explanatory {\em Power}.

This endeavour has been largely successful: much of the SP system is the relatively simple but powerful concept of SP-multiple-alignment (described in Appendix \ref{sp-multiple-alignments_appendix}) which is largely responsible for the system's versatility in aspects of intelligence (Section \ref{versatility_in_ai_section}), including several kinds of reasoning (Section \ref{versatility_in_reasoning_section}), its versatility in the representation of diverse kinds of knowledge (Section \ref{versatility_in_rk_section}), and its potential for the seamless integration of diverse aspects of intelligence and diverse kinds of knowledge, in any combination (Section \ref{seamless_integration_section}).

The SP system, with SP-multiple-alignment at centre stage, provides a much more favourable combination of ``Simplicity'' with ``Power'' than any alternatives, including the kinds of CSRK-related systems outlined in Section \ref{research_on_cskr_section}, and the many attempts to develop `unified theories of cognition' (see, for example, \cite{newell_1992,newell_1990}) and `artificial general intelligence' (see, for example, \cite{goertzel_pennachin_2007}).

This is not some meaningless abstraction. Amongst other things it means that, to the extent that the SP system provides solutions to problems in CSRK, it is likely that those solutions will integrate smoothly amongst themselves and with other kinds of information processing, something that appears to be essential in any theory of CSRK or, more generally, human intelligence (Section \ref{seamless_integration_section}).

In this context, it is relevant to note that, since the SP system and the concept of SP-multiple-alignment have been developed to simplify and integrate observations and concepts across a broad canvass (Appendix \ref{distinctive_features_and_advantages_appendix}), it will of course have points of resemblance to many other systems. But any such resemblance does not mean that the SP system is ``nothing but X'', or ``nothing but Y'', and should not be a distraction from the importance of simplification and integration in IT systems, and the relative success of the SP system in combining conceptual simplicity with high levels of descriptive and explanatory power.

\subsection{Turing Equivalence, Plus Aspects of Human Intelligence}

The SP system is, probably, Turing-equivalent in the sense that it can in principle perform any computation within the scope of a universal Turing machine \cite[Chapter 4]{wolff_2006}. But, unlike a `raw' Turing machine (without any programming) or a `raw' conventional computer, it has strengths and potential in AI (\cite{sp_extended_overview,wolff_2006}), something which, as Turing recognised (\cite{turing_1950,webster_2012}), is missing from the universal Turing machine. Thus the SP system has the kind of generality needed for CSRK, and its strengths and potential in AI give it a head start in modelling CSRK.

\subsection{Information Compression}

As described in Appendix \ref{organisation_and_workings_appendix}, a central part of the SP system is `information compression via the matching and unification of patterns' (ICMUP) and, more specifically, `information compression via SP-multiple-alignment' (ICSPMA).

For CSRK, information compression in the SP system has a three-fold significance:

\begin{itemize}

    \item {\em Generality in inference and probability}. As noted in Appendix \ref{origins_development_appendix}, the intimate connection that is known to exist between information compression and concepts of inference and probability means that the SP system has strengths and potential in inference and in the calculation of probabiliies.

    \item {\em Generality in the representation of knowledge}. The generality of information compression suggests that, in principle, {\em any} kind of knowledge may be represented in a succinct form in the SP system.

    \item {\em The DONSVIC principle}. Unsupervised learning in the SP Computer Model as it has been developed to date, conforms to the DONSVIC principle \cite[Section 5.2]{sp_extended_overview}, meaning that knowledge structures created by the system are, generally, ones that people regard as natural and which yield relatively high levels of information compression. It appears that such forms of knowledge are those that are most relevant to CSK.

\end{itemize}

\subsection{Versatility in Aspects of Intelligence}\label{versatility_in_ai_section}

Largely because of the mechanisms of SP-multiple-alignment, the SP system demonstrates strengths and potential in several aspects of intelligence including: unsupervised learning; natural language processing; fuzzy pattern recognition; recognition at multiple levels of abstraction; best-match and semantic forms of information retrieval; several kinds of reasoning (more in Section \ref{versatility_in_reasoning_section}, below); planning; and problem solving (\cite[Chapters 5 to 9]{wolff_2006}, \cite{sp_extended_overview}).

Subsections that follow, describe a selection of these aspects of intelligence, and how they may be modelled with the SP Computer Model.

\subsubsection{Pattern Recognition}\label{pattern_recognition_etc_section}

Figure \ref{class_part_plant_figure} shows how, via the building of an SP-multiple-alignment, the SP computer system may model the recognition of an unknown plant.

The first thing to make clear is that, compared with the SP-multiple-alignment in Figure \ref{fruit_flies_figure}, this SP-multiple-alignment is rotated by $90\degree$, with SP-patterns arranged in columns instead of rows and alignments between matching symbols shown in rows instead of columns. The choice between these two ways of displaying SP-multiple-alignments depends purely on what fits best on the page.

\begin{figure}[!htbp]
\fontsize{06.00pt}{07.20pt}
\centering
{\bf
\begin{BVerbatim}
0                 1                2                  3              4              5                  6

                  <species>
                  acris
                  <genus> ---------------------------------------------------------------------------- <genus>
                  Ranunculus ------------------------------------------------------------------------- Ranunculus
                                                                                    <family> --------- <family>
                                                                                    Ranunculaceae ---- Ranunculaceae
                                                                     <order> ------ <order>
                                                                     Ranunculales - Ranunculales
                                                      <class> ------ <class>
                                                      Angiospermae - Angiospermae
                                   <phylum> --------- <phylum>
                                   Plants ----------- Plants
                                   <feeding>
has_chlorophyll ------------------ has_chlorophyll
                                   photosynthesises
                                   <feeding>
                                   <structure> ------ <structure>
                                                      <shoot>
<stem> ---------- <stem> ---------------------------- <stem>
hairy ----------- hairy
</stem> --------- </stem> --------------------------- </stem>
                  <leaves> -------------------------- <leaves>
                  compound
                  palmately_cut
                  </leaves> ------------------------- </leaves>
                                                      <flowers> ------------------- <flowers>
                                                                                    <arrangement>
                                                                                    regular
                                                                                    all_parts_free
                                                                                    </arrangement>
                  <sepals> -------------------------------------------------------- <sepals>
                  not_reflexed
                  </sepals> ------------------------------------------------------- </sepals>
<petals> -------- <petals> -------------------------------------------------------- <petals> --------- <petals>
                                                                                    <number> --------- <number>
                                                                                                       five
                                                                                    </number> -------- </number>
                  <colour> -------------------------------------------------------- <colour>
yellow ---------- yellow
                  </colour> ------------------------------------------------------- </colour>
</petals> ------- </petals> ------------------------------------------------------- </petals> -------- </petals>
                                                                                    <hermaphrodite>
<stamens> ------------------------------------------------------------------------- <stamens>
numerous -------------------------------------------------------------------------- numerous
</stamens> ------------------------------------------------------------------------ </stamens>
                                                                                    <pistil>
                                                                                    ovary
                                                                                    style
                                                                                    stigma
                                                                                    </pistil>
                                                                                    </hermaphrodite>
                                                      </flowers> ------------------ </flowers>
                                                      </shoot>
                                                      <root>
                                                      </root>
                                   </structure> ----- </structure>
<habitat> ------- <habitat> ------ <habitat>
meadows --------- meadows
</habitat> ------ </habitat> ----- </habitat>
                  <common_name> -- <common_name>
                  Meadow
                  Buttercup
                  </common_name> - </common_name>
                                   <food_value> ----------------------------------- <food_value>
                                                                                    poisonous
                                   </food_value> ---------------------------------- </food_value>
                                   </phylum> -------- </phylum>
                                                      </class> ----- </class>
                                                                     </order> ----- </order>
                                                                                    </family> -------- </family>
                  </genus> --------------------------------------------------------------------------- </genus>
                  </species>

0                 1                2                  3              4              5                  6
\end{BVerbatim}
}
\caption{The best SP-multiple-alignment created by the SP Computer Model, with a small set of New SP-patterns (in column 0) that describe some features of an unknown plant, and a set of Old SP-patterns, including those shown in columns 1 to 6, that describe different categories of plant, with their parts and sub-parts, and other attributes. Reproduced with permission from Figure 16 in \protect\cite{sp_extended_overview}.}
\label{class_part_plant_figure}
\end{figure}

For the creation of the SP-multiple-alignment shown in Figure \ref{class_part_plant_figure}, the SP Computer Model was supplied with:

\begin{itemize}

    \item {\raggedright Five New SP-patterns, shown in column 0 of the figure: the one-symbol SP-pattern `\texttt{has\_chlorophyll}', and four multi-symbol SP-patterns, `\texttt{<stem> hairy </stem>}', `\texttt{<petals> yellow </petals>}', `\texttt{<stamens> numerous </stamens>}', and `\texttt{<habitat> meadows </habitat>}', which describe the features of some unknown plant}. These New SP-patterns may be supplied to the SP Computer Model in any order, not only the order shown in column 0 of the figure.

    \item A relatively large set of Old SP-patterns including those shown in columns 1 to 6. These describe the structures and attributes of different kinds of plant.

\end{itemize}

As with the example discussed in Appendix \ref{building_sp-multiple-alignments_appendix}, the SP system tries to find one or more SP-multiple-alignments where, in each one, the New SP-pattern or SP-patterns may be encoded economically in terms of the Old SP-patterns in the given SP-multiple-alignment. Broadly speaking, this means finding plenty of matches between New SP-patterns and Old SP-patterns, and a good number of matches between Old SP-patterns.

In the same way that the New SP-pattern or SP-patterns are always shown in row 0 of `horizontally' arranged SP-multiple-alignments like those shown in Figure \ref{fruit_flies_figure}, with Old SP-patterns, one per row, in the remaining rows, the New SP-pattern or SP-patterns of `vertically' arranged SP-multiple-alignments like that shown in Figure \ref{class_part_plant_figure} are, by convention, always shown in column 0, and the Old SP-patterns are shown in the remaining columns, one SP-pattern per column. Otherwise, the order of SP-patterns across the columns has no significance.

As described in Appendix \ref{building_sp-multiple-alignments_appendix}, the aim in creating SP-multiple-alignments is to find ones which are `good' in terms of the economical encoding of the New SP-pattern(s) in terms of the Old SP-patterns in that SP-multiple-alignment. In the process of searching for such SP-multiple-alignments, the SP system creates many SP-multiple-alignments or partial SP-multiple-alignments and discards most of them. Eventually, it is left with a few SP-multiple-alignments which are good, often as few as one or two.

The SP-multiple-alignment in Figure \ref{class_part_plant_figure} is the best of those created by the SP Computer Model. It identifies the unknown plant with the SP-pattern shown in column 1: it is the species {\em acris} and it has the common name `Meadow Buttercup'.

\subsubsection{Class-Inclusion Relations and Part-Whole Relations}\label{class-inclusion_and_part-whole_relations_section} 

A feature of Figure \ref{class_part_plant_figure} that has not so far been mentioned is that the unknown plant is not only identified as the species {\em acris} with the common name `Meadow Buttercup' (column 1) but it is also recognised as belonging to the genus {\em Ranunculus} (column 6). And the unknown plant is also recognised as belonging to the family Ranunculaceae (column 5), which is in the order Ranunculales (column 4), in the class Angiospermae (column 3), and in the phylum Plants (column 2).

In short, the SP system provides for the representation of class-inclusion hierarchies, and for their being an integral part of the recognition process, providing for recognition at multiple levels of abstraction, as mentioned near the beginning of Section \ref{versatility_in_ai_section}. Notice how different New SP-symbols may be matched with Old SP-symbols at different levels in the class hierarchy: the feature `\texttt{<stamens> numerous </stamens>}' is matched at the `\texttt{family}' level (column 5), the feature `\texttt{has\_chlorophyl}' is matched at the `\texttt{Plants}' level (column 2), the feature `\texttt{<stem> hairy </stem>}' is matched at the `\texttt{species}' level (column 1), and so on.

A second important feature of Figure \ref{class_part_plant_figure} is that it shows how the SP system not only supports the representation of class-inclusion hierarchies but it also supports the representation of part-whole hierarchies. For example, attributes which have a part-whole hierarchical structure include: `\texttt{flowers}' in column 3; broken down into `\texttt{sepals}', `\texttt{petals}', `\texttt{stamens}', and other attributes in column 5, with a further breakdown of `\texttt{petals}' into the attributes `\texttt{<number> </number>}' and `\texttt{<colour> </colour>}'. Actual values for the latter two attributes are, in this example, `\texttt{<number> five </number>}' in column 6, and `\texttt{<colour> yellow </colour>}' in column 1.

The figure illustrates an important feature of the SP system: that there can be seamless integration of class-inclusion relations with part-whole relations, as discussed under `Categories and properties' in  Section \ref{taxonomic_reasoning_section} below.

\subsubsection{Inheritance of Attributes}\label{inheritance_of_attributes_section}

An important aspect of recognition in the SP system with class-inclusion hierarchies and with part-whole hierarchies is that both kinds of hierarchy provide a means of making a type of inference called ``inheritance of attributes'' that is bread-and-butter in everyday reasoning and everyday thinking.\footnote{It relates to ``Prediction by partial matching'' in information compression (see, for example, \cite{teahan_alhawiti_2015}) which means predicting the unseen parts of a pattern that has been recognised.}

To see inheritance of attributes in action, we may infer from the SP-multiple-alignment shown in Figure \ref{class_part_plant_figure} that, in the plant that has been identified as {\em Ranunculus acris}, there are sepals that are not reflexed and leaves that are compound and palmately cut (the `\texttt{species}' level in column 1), that the plant nourishes itself via photosynthesis (the `\texttt{phylum}' level in column 2), and that it is poisonous (the `\texttt{family}' level in column 5). If there was more detail in the SP-patterns in the example, many more such inferences would be possible.

The intimate relation between pattern recognition and inference via inheritance of attributes illustrates a general truth about the SP system: that there is potential for the seamless integration of different aspects of intelligence, as discussed in Section \ref{seamless_integration_section}, below.

\subsubsection{Recognition in the Face of Errors of Omission, Commission, or Substitution}\label{recognition_despite_errors_section}

\begin{figure}[!htbp]
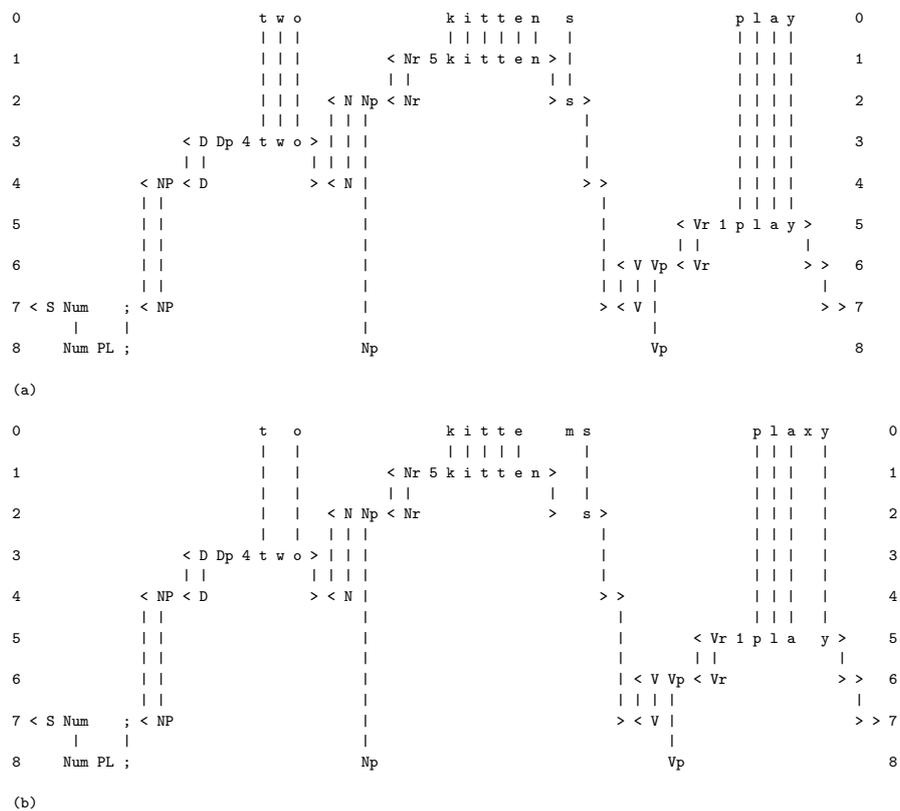

\fontsize{06.50pt}{07.80pt}
\centering
{\bf
\begin{BVerbatim}
0                            t w o                 k i t t e n   s                   p l a y       0
                             | | |                 | | | | | |   |                   | | | |
1                            | | |          < Nr 5 k i t t e n > |                   | | | |       1
                             | | |          | |                | |                   | | | |
2                            | | |   < N Np < Nr               > s >                 | | | |       2
                             | | |   | | |                         |                 | | | |
3                   < D Dp 4 t w o > | | |                         |                 | | | |       3
                    | |            | | | |                         |                 | | | |
4              < NP < D            > < N |                         > >               | | | |       4
               | |                       |                           |               | | | |
5              | |                       |                           |        < Vr 1 p l a y >     5
               | |                       |                           |        | |            |
6              | |                       |                           | < V Vp < Vr           > >   6
               | |                       |                           | | | |                   |
7 < S Num    ; < NP                      |                           > < V |                   > > 7
       |     |                           |                                 |
8     Num PL ;                           Np                                Vp                      8

(a)

0                            t   o                 k i t t e     m s                   p l a x y       0
                             |   |                 | | | | |       |                   | | |   |
1                            |   |          < Nr 5 k i t t e n >   |                   | | |   |       1
                             |   |          | |                |   |                   | | |   |
2                            |   |   < N Np < Nr               >   s >                 | | |   |       2
                             |   |   | | |                           |                 | | |   |
3                   < D Dp 4 t w o > | | |                           |                 | | |   |       3
                    | |            | | | |                           |                 | | |   |
4              < NP < D            > < N |                           > >               | | |   |       4
               | |                       |                             |               | | |   |
5              | |                       |                             |        < Vr 1 p l a   y >     5
               | |                       |                             |        | |              |
6              | |                       |                             | < V Vp < Vr             > >   6
               | |                       |                             | | | |                     |
7 < S Num    ; < NP                      |                             > < V |                     > > 7
       |     |                           |                                   |
8     Num PL ;                           Np                                  Vp                        8

(b)
\end{BVerbatim}
}
\caption{(a) The best SP-multiple-alignment created by the SP Computer Model with a store of Old SP-patterns like those in rows 1 to 8 (representing grammatical structures, including words) and a New SP-pattern (representing a sentence to be parsed) shown in row 0. (b) As in (a) but with errors of omission, commission and substitution, and with same set of Old SP-patterns as before. (a) and (b) are reproduced from Figures 1 and 2 in \protect\cite{wolff_sp_intelligent_database}, with permission.}
\label{noisy_data_recognition_figure}
\end{figure}

An aspect of pattern reccognition via the SP system that is not illustrated in Figure \ref{class_part_plant_figure} is that, like people, the system has a robust ability to recognise patterns despite errors of omission, commission, and substitution in the pattern or patterns that are to be recognised. An example of this aspect of pattern recognition, mentioned as ``fuzzy'' pattern recognition near the beginning of Section \ref{versatility_in_ai_section}, is shown in Figure \ref{noisy_data_recognition_figure} (b).

This examples shows how, compared with Figure \ref{noisy_data_recognition_figure} (a), the SP system may achieve a `correct' parsing of the sentence `\texttt{t w o k i t t e n s p l a y}' despite errors of omission (like `\texttt{t o}' instead of `\texttt{t w o}' in the New pattern), errors of commission (like `\texttt{p l a x y}' instead of `\texttt{p l a y}'), and errors of substitution (like `\texttt{k i t t e m s}' instead of `\texttt{k i t t e n s}').

\subsection{Versatility in Kinds of Reasoning}\label{versatility_in_reasoning_section}

Although reasoning is an aspect of intelligence (Section \ref{versatility_in_ai_section}), it has been given a subsection to itself because of the versatility of the SP system in this area and because of its potential with CSR.

In reasoning, strengths and potential of the SP system, described quite fully in \cite[Chapter 7]{wolff_2006} and \cite[Section 10]{sp_extended_overview}, are in brief: one-step `deductive' reasoning; chains of reasoning; abductive reasoning; reasoning with probabilistic networks and trees; reasoning with `rules'; nonmonotonic reasoning and reasoning with default values; Bayesian reasoning with `explaining away' (as discussed by Judea Pearl in \cite[Sections 1.2.2 and 2.2.4]{pearl_1997}); causal reasoning; and reasoning that is not supported by evidence.

As we have seen in Section \ref{inheritance_of_attributes_section}, the SP system also supports inference via inheritance of attributes. It appears that there is also potential for spatial reasoning \cite[Section IV-F.1]{sp_autonomous_robots}, and for what-if reasoning \cite[Section IV-F.2]{sp_autonomous_robots}.

It would not be either feasible or appropriate to reproduce everything in this area that has been published before. But to see some of the versatility of the SP system in modelling different kinds of reasoning---and corresponding potential in CSR---readers are urged to look at the afore-mentioned sources, \cite[Chapter 7]{wolff_2006} and \cite[Section 10]{sp_extended_overview}, and, in particular:

\begin{itemize}

    \item How the SP system may model nonmonotonic reasoning and reasoning with default values (\cite[Section 10.1]{sp_extended_overview}, \cite[Section 7.7]{wolff_2006}).

    \item How the SP system may model Bayesian networks with `explaining away' (\cite[Section 10.2]{sp_extended_overview}, \cite[Section 7.8]{wolff_2006}).

\end{itemize}

\subsection{Versatility in the Representation of Knowledge}\label{versatility_in_rk_section}

The quest for simplification and integration of observations and concepts in AI and related areas (Appendix \ref{distinctive_features_and_advantages_appendix}) has led to the creation of a system that combines simplicity in its organisation with versatility in diverse aspects of intelligence (Section \ref{versatility_in_ai_section}), including versatility in reasoning (Section \ref{versatility_in_reasoning_section}). This section outlines how the SP system exhibits versatility in the representation of diverse forms of knowledge.

As noted in Appendix \ref{outline_of_sp_system_appendix}, it is envisaged that all kinds of knowledge in the SP system are to be represented with {\em SP-patterns} meaning arrays of atomic SP-symbols in one or two dimensions.

Despite their simplicity, SP-patterns, within the SP-multiple-alignment framework, have strengths and potential in representing several forms of knowledge, any of which may serve in CSK. These include: the syntax of natural language (see Figure \ref{fruit_flies_figure}); class hierarchies (see Figure \ref{class_part_plant_figure}), class heterarchies (meaning class hierarchies with cross-classification); part-whole hierarchies (see Figure \ref{class_part_plant_figure}); discrimination networks and trees; entity-relationship structures; relational knowledge; rules for use in reasoning; SP-patterns in two dimensions; images; structures in three dimensions; and procedural knowledge.

There is more detail throughout \cite{wolff_2006} and \cite{sp_extended_overview}, and there are references to further sources of information in \cite[Section III-B]{sp_big_data}. Some examples have been seen in Figure \ref{class_part_plant_figure} and more are shown later.

\subsection{Seamless Integration of Diverse Aspects of Intelligence and Diverse Forms of Knowledge in Any Combination}\label{seamless_integration_section}

Three features of the SP system suggests that it should facilitate seamless integration of diverse aspects of intelligence (including several forms of reasoning) and seamless integration of diverse kinds of knowledge, in any combination:

\begin{itemize}

    \item The adoption of one simple format---SP-patterns---for all kinds of knowledge;

    \item That one relatively simple framework---SP-multiple-alignment---is central in all kinds of processing;

    \item That the relatively simple format for knowledge and the SP-multiple-alignment framework provides for several aspects of intelligence (Section \ref{versatility_in_ai_section}) including several kinds of reasoning (Section \ref{versatility_in_reasoning_section}), and for the representation of several different kinds of knowledge (Section \ref{versatility_in_rk_section}).

\end{itemize}

Seamless integration of structures and processes may be seen in the examples relating to Figure \ref{class_part_plant_figure}:

\begin{itemize}

    \item Class-inclusion relations and part-whole relations work together in the representation of knowledge without awkward incompatibilities (Section \ref{class-inclusion_and_part-whole_relations_section}).

    \item Pattern recognition (an aspect of general intelligence) and inheritance of attributes (a form of reasoning) are intimately related in what is, in effect, one type of operation (Sections \ref{class-inclusion_and_part-whole_relations_section} and \ref{inheritance_of_attributes_section}).

\end{itemize}

For the understanding of natural language and the production of language from meanings, it is likely to be helpful if there is seamless integration of syntax and semantics, and it seems likely that this will be facilitated by representing both of them with SP-patterns, and by processing both of them together via the building and manipulation of SP-multiple-alignments.

Some preliminary examples from the SP Computer Model show how this kind of integration may be achieved with both the understanding and production of natural language \cite[Section 5.7]{wolff_2006}. There is clear potential with the SP system for the comprehensive integration of the syntax and semantics of natural language.

Seamless integration of diverse aspects of intelligence and diverse kinds of knowledge in any combination is likely to be critically important in the modelling of CSRK since, as a matter of ordinary experience, CSR means a willingness to use any and all relevant forms of knowledge, and a willingness to be flexible in one's thinking, using diverse forms of reasoning with diverse kinds of intelligence where appropriate. More generally, that kind of seamless integration appears to be {\em essential} in any artificial system that aspires to the fluidity, versatility, and adaptability of human intelligence.

\section{Aspects of Commonsense Reasoning and Commonsense Knowledge and How They May Be Modelled in the SP System}\label{aspects_of_csrk_section}

This main section discusses aspects of CSRK in the light of what has been said about the SP system earlier in this paper and in Appendix \ref{outline_of_sp_system_appendix}. The discussion focuses mainly on what DM have said about CSRK, and uses several of the headings in that paper.

\subsection{Father and Son, and Other Examples}\label{father_son_etc_section}

As was mentioned in the Introduction, DM give some examples of sentences that describe everyday situations that are easy for people to understand but can be difficult for AI systems:

\begin{itemize}

    \item ``Who is taller, Prince William or his baby son Prince George?'' DM say ``... if you see a six-foot-tall person holding a two-foot-tall person in his arms, and you are told they are father and son, you do not have to ask which is which.'' (p.~92).

    \item ``Can you make a salad out of a polyester shirt?'' DM say ``If you need to make a salad for dinner and are out of lettuce, you do not waste time considering improvising by taking a shirt [out] of the closet and cutting it up.'' (pp.~92--93).

    \item DM say ``If you read the text, `I stuck a pin in a carrot; when I pulled the pin out, it had a hole,' you need not consider the possibility `it' refers to the pin.'' (p.~93).

\end{itemize}

Although preliminary work, mentioned in Section \ref{seamless_integration_section}, shows how, in the SP system, syntactic and semantic knowledge can work together in the understanding of natural language, more work would be needed to demonstrate an understanding of example sentences like those just shown.

But the kinds of inferences needed for the understanding of those sentences are well within the scope of the SP system. It appears that all three of them depend largely on inheritance of attributes, discussed in Section \ref{inheritance_of_attributes_section}:

\begin{itemize}

    \item With the father and son example, height is the kind of attribute that would normally be associated with people of all kinds and with subclasses like `father' and `son'. And that knowledge would suggest immediately, via inheritance of attributes, that the father would be taller than the son, especially since the son is described as a baby.

        Although the inference is probabilistic, the information that Prince George is a baby, the knowledge that most people have about Prince
        William, and the difficulty that a small person would have in holding a big person in their arms, would, very likely, rule out the
        possibility that the father is a dwarf and that the son is fully grown.

    \item The salad example depends more directly on inheritance of attributes: anything that goes into a salad must have the attribute `edible', at any level in the hierarchy or hierarchies of classes in which it belongs. A polyester shirt clearly fails that test. This is illustrated in the SP-multiple-alignment shown in Figure \ref{salad_figure}, as discussed below.

    \item In a similar way, the carrot and pin example depends, at least in part, on characteristics of carrots (they are relatively soft) and pins (they are normally made of something that is relatively hard, normally metal, and they are normally designed to stick into things) that may be inherited from any of the levels in the classes in which they belong. Also relevant is the meanings of the words in the phrase ``stuck X in Y'', which implies that X would make a hole in Y.

\end{itemize}

In Figure \ref{salad_figure}, column 0 contains a New SP-pattern with just one SP-symbol: `\texttt{salad}'. The remaining columns show some of the Old SP-patterns supplied to the SP Computer Model, one SP-pattern per column. From this SP-multiple-alignment, we may see that the dish is classified as a salad (column 1), that it is `\texttt{savoury}' (column 2), and that it is a type of `\texttt{dish}' (column 3). As a dish, it contains a list of ingredients represented with the recursive SP-pattern `\texttt{<ingredients> ig1 <ingrdnt> edible </ingrdnt> <ingredients> </ingredients> </ingredients>}'.

Since recursion in the SP system may not be familiar to readers, a word of explanation is given here. The first point to clarify is that any Old SP-pattern may appear more than once in any SP-multiple-alignment, even though each of the instances is not a copy of any other. In accordance with that principle, the SP-pattern mentioned at the end of the preceding paragraph---`\texttt{<ingredients> ig1 <ingrdnt> edible </ingrdnt> <ingredients> </ingredients> </ingredients>}'---appears 3 times in the SP-multiple-alignment in Figure \ref{salad_figure}, in columns 4, 5, and 8.

The reason that it can appear 3 times is that, although any given SP-symbol cannot be aligned with itself within an SP-multiple-alignment, the SP-symbol `\texttt{<ingredients>}' at the beginning of the SP-pattern can be aligned with a copy of that SP-symbol which appears at the 6th position in the SP-pattern, and the SP-symbol `\texttt{</ingredients>}', which appears at the 7th position, can be aligned with a copy of that SP-symbol which appears at the 8th position in the SP-pattern. As with any kind of recursion, the repeated structure can in principle appear any number of times.

\newgeometry{left=1cm,right=1cm}

\begin{figure}[!htbp]
\fontsize{06.00pt}{07.20pt}
\centering
{\bf
\begin{BVerbatim}
0       1            2            3                4                5                6            7            8                9

        <salad>
        sld1
        <savoury> -- <savoury>
                     sv1
                     savoury
                     <dish> ----- <dish>
                                  dsh1
        <name> ----- <name> ----- <name>
salad - salad
        </name> ---- </name> ---- </name>
                                  <ingredients> -- <ingredients>
                                                   ig1
                                                   <ingrdnt> ------------------------------------ <ingrdnt>
                                                                                                  fd4
                                                   edible --------------------------------------- edible
                                                                                                  cucumber
                                                   </ingrdnt> ----------------------------------- </ingrdnt>
                                                   <ingredients> -- <ingredients>
                                                                    ig1
                                                                    <ingrdnt> ------ <ingrdnt>
                                                                                     fd5
                                                                    edible --------- edible
                                                                                     radish
                                                                    </ingrdnt> ----- </ingrdnt>
                                                                    <ingredients> ---------------------------- <ingredients>
                                                                                                               ig1
                                                                                                               <ingrdnt> ------ <ingrdnt>
                                                                                                                                fd3
                                                                                                               edible --------- edible
                                                                                                                                lettuce
                                                                                                               </ingrdnt> ----- </ingrdnt>
                                                                                                               <ingredients>
                                                                                                               </ingredients>
                                                                    </ingredients> --------------------------- </ingredients>
                                                   </ingredients> - </ingredients>
                                  </ingredients> - </ingredients>
                     </dish> ---- </dish>
        </savoury> - </savoury>
        </salad>

0       1            2            3                4                5                6            7            8                9
\end{BVerbatim}
}
\caption{An SP-multiple-alignment from the SP Computer Model showing how a `\texttt{salad}' may inherit the feature `\texttt{edible}' from all its ingredients.}
\label{salad_figure}
\end{figure}

\restoregeometry

The key point for present purposes is that every one of the ingredients---cucumber, radish, and lettuce in this example---is marked as `\texttt{edible}', and likewise for all the other Old SP-patterns supplied to the model. A polyester shirt would not appear anywhere amongst the ingredients of any salad dish.

A more fully-developed version of this example would contain controls on the kinds of ingredients that may go into each type of dish (it would, for example, be somewhat eccentric to put ice cream in a salad) and controls to ensure that, normally, for any given dish, each type of ingredient would appear only once.

\subsection{The City Councilmen Refused the Demonstrators a Permit}\label{city_councilmen_section}

As noted by DM (p.~93), the subtlety of natural language may be seen in a pair of example sentences presented by Terry Winograd in \cite[p.~33]{winograd_1972}:

\begin{itemize}

    \item {\em The city councilmen refused the demonstrators a permit because they feared violence}.

    \item {\em The city councilmen refused the demonstrators a permit because they advocated revolution}.

\end{itemize}

People naturally assume that, in the first sentence, ``they'' means the city councilmen, while in the second sentence, ``they'' means the demonstrators, but those two inferences may be problematic for AI systems.

A relatively full interpretation of sentences like these will be presented in \cite{sp_syntax_semantics_2018}. For now, the key to the interpretation of these sentences, appears to be inheritance of attributes as discussed in Section \ref{inheritance_of_attributes_section}.

In this case, inheritance of attributes may work something like this: it is generally known that, as a class, city councilmen will normally wish to maintain peace in their city, while ``demonstrators'' do sometimes advocate revolution, and revolutions are often violent. Inferences from that knowledge (via inheritance of attributes), with a knowledge of the meanings of the words ``feared'' and ``advocated'', appears to be sufficient, in each of the two cases above, to disambiguate the pronoun ``they''.

\subsection{Computer Vision}\label{computer_vision_section}

With respect to a photograph of a kitchen shown in \cite[Figure 1]{davis_marcus_2015}, DM say:

\begin{quote}

    ``Many of the objects that are small or partially seen, such as the metal bowls in the shelf on the left, the cold water knob for the faucet, the round metal knobs on the cabinets, the dishwasher, and the chairs at the table seen from the side, are only recognizable in context; the isolated image would be difficult to identify.'' (p.~94).

\end{quote}

\noindent They go on to say that ``The viewer infers the existence of objects that are not in the image at all.'' \cite[p.~94]{davis_marcus_2015} and that ``The viewer also infers how the objects can be used (sometimes called their `affordances');'' ({\em ibid.}).

These kinds of capabilities appear to be well within the scope of the SP system, as described in the next two subsections.

\subsubsection{The Importance of Context in Recognition}

Although the example to be discussed here is from the processing of natural language, it should be clear that the principles may also apply to vision. Consider the verbal expression `\texttt{ae i s k r ee m}', written with a simplified notation for speech sounds or `phonemes'. This may be read as ``I scream'' or ``ice cream''. As in the example shown in Figure \ref{fruit_flies_figure}, modelling that kind of ambiguity is well within the scope of the SP Computer Model, as shown in Figure \ref{icecream_alignment_1}.

\begin{figure}[!hbt]
\fontsize{12.00pt}{14.40pt}
\centering
\begin{BVerbatim}
0          ae i s        k r ee m        0
           |  | |        | | |  |
1          |  | |    N 0 k r ee m #N     1
           |  | |    |            |
2 NP 2 A   |  | | #A N            #N #NP 2
       |   |  | | |
3      A 0 ae i s #A                     3

(a)

0          ae i         s k r ee m                0
           |  |         | | | |  |
1          |  |     V 0 s k r ee m #V             1
           |  |     |              |
2 S 0 NP   |  | #NP V              #V ADV #ADV #S 2
      |    |  |  |
3     NP 1 ae i #NP                               3

(b)
\end{BVerbatim}
\caption{The two best parsings of the phoneme sequence `\texttt{ae i s k r ee m}' created by the SP Computer Model. Reproduced with permission from Figure 5.3 in \cite{wolff_2006}.}
\label{icecream_alignment_1}
\end{figure}

In accordance with ordinary experience, that kind of ambiguity in speech or in vision may be reduced or eliminated by introducing disambiguating contexts. The way in which the SP system may model that effect is shown in Figure \ref{icecream_alignment_2}.

\begin{figure}[!hbt]
\fontsize{09.00pt}{10.80pt}
\centering
\begin{BVerbatim}
0          ae i         s k r ee m          l ae w d l i         0
           |  |         | | | |  |          | |  | | | |
1          |  |         | | | |  |    ADV 0 l ae w d l i #ADV    1
           |  |         | | | |  |     |                  |
2 S 0 NP   |  | #NP V   | | | |  | #V ADV                #ADV #S 2
      |    |  |  |  |   | | | |  | |
3     |    |  |  |  V 0 s k r ee m #V                            3
      |    |  |  |
4     NP 1 ae i #NP                                              4

(a)

0              ae i s        k r ee m             i z         k o l d       0
               |  | |        | | |  |             | |         | | | |
1              |  | |        | | |  |             | |     A 2 k o l d #A    1
               |  | |        | | |  |             | |     |           |
2 S 1 NP       |  | |        | | |  |    #NP VB   | | #VB A           #A #S 2
      |        |  | |        | | |  |     |  |    | |  |
3     |        |  | |    N 0 k r ee m #N  |  |    | |  |                    3
      |        |  | |    |            |   |  |    | |  |
4     NP 2 A   |  | | #A N            #N #NP |    | |  |                    4
           |   |  | | |                      |    | |  |
5          A 0 ae i s #A                     |    | |  |                    5
                                             |    | |  |
6                                            VB 0 i z #VB                   6

(b)
\end{BVerbatim}
\caption{(a) The best parsing found by the SP Computer Model for the phoneme sequence `ae i s k r ee m' when it is contained within the larger sequence `ae i s k r ee m l ae w d l i'. (b) The best parsing of the phoneme sequence `ae i s k r ee m' when it is included within the larger sequence `ae i s k r ee m i z k o l d'. Reproduced with permission from Figure 5.4 in \cite{wolff_2006}.}
\label{icecream_alignment_2}
\end{figure}

In Figure \ref{icecream_alignment_2} (a), the sequence `\texttt{ae i s k r ee m}' is contained within the larger sequence `\texttt{ae i s k r ee m l ae w d l i}'. In this case, there is no ambiguity in the parsing: there is just one best result corresponding to the expression ``I scream loudly''.

In Figure \ref{icecream_alignment_2} (b), the sequence `\texttt{ae i s k r ee m}' is contained within the larger sequence `\texttt{ae i s k r ee m i z k o l d}'. In this case, as before, there is no ambiguity in the parsing: there is just one best result corresponding to the expression ``ice cream is cold''.

\subsubsection{`Seeing' Things That Are Not Objectively Present in an Image}

The SP Computer Model provides an account of some types of situation where people `see' things in an image that are not objectively present in that image:

\begin{itemize}

    \item {\em Inference of missing entities}. In the example of recognition illustrated in Figure \ref{class_part_plant_figure}, we may infer that the unknown plant has features that were not in the New information supplied to the SP Computer Model. Examples of such features in Figure \ref{class_part_plant_figure} include sepals that are not reflexed, leaves that are compound and palmately cut, that the plant nourishes itself via photosynthesis, and that it is poisonous. In a similar way, in computer vision, we may infer objects that are not shown explicitly in an image, and uses for an object that are not visible either.

    \item {\em Inference of missing boundaries between entities}. David Marr \cite{marr_2010} describes two examples where people `see' things that are not objectively present in an image: the sides of a triangle that can be seen in Kanizsa's triangle although it is mostly empty space ({\em ibid.}, Figure 2-6); and the line in a photograph of a plant, where one leaf overlaps another leaf, that we can `see' although there is nothing objective to mark it ({\em ibid.}, Figure 4-1 (a)). These things may be seen to be the result of a process of visual parsing that has the effect of introducing boundaries between segments, although those boundaries are not objectively present in what is being parsed \cite[Section 7.1]{sp_vision}.

\end{itemize}

Figure \ref{noisy_data_recognition_figure} (a) illustrates the inference of missing boundaries between entities. In the parsing, `\texttt{t w o}' and `\texttt{k i t t e n}' are identified as discrete entities with a boundary between them which is not objectively present in the raw data. Likewise for the rest of that parsing, and the parsing examples shown in Figures \ref{icecream_alignment_1} and \ref{icecream_alignment_2}.

Another example showing how the SP Computer Model may discover missing boundaries is \cite[Section XI-A.2, Figure 8]{sp_alternatives} which shows how, via parsing, the invisible boundaries in Kanizsa's triangle may be identified.

\subsection{Robotic Manipulation}\label{robotic_manipulation_section}

Although the SP concepts appear to be highly relevant to the development of intelligence in robots, as described in \cite{sp_autonomous_robots}, no attempt has yet been made to address problems that DM describe like this:

\begin{quote}

    ``If a cat runs in front of a house-cleaning robot, the robot should neither run it over nor sweep it up nor put
    it away on a shelf. These things seem obvious, but ensuring a robot avoids mistakes of this kind is very
    challenging.'' \cite[p.~94]{davis_marcus_2015}.

\end{quote}

\noindent That said, it seems likely that the SP system's strengths and potential in diverse forms of reasoning (Section \ref{versatility_in_reasoning_section}, \cite[Section IV-F]{sp_autonomous_robots}) will help it to avoid the kinds of mistake sketched by DM.

\subsection{Successes in Automated Commonsense Reasoning}\label{successes_in_csr_section}

This section considers four areas where, as described by DM, there have been successes in CSRK-related processing \cite[pp.~94--97]{davis_marcus_2015}. These four approaches, and what DM have to say about them, are discussed here in relation to the SP system and what it can do.

\subsubsection{Taxonomic Reasoning}\label{taxonomic_reasoning_section}

The first area of success in CSRK-related processing discussed by DM is taxonomic reasoning. They say ``Simple taxonomic structures such
as those illustrated [in \cite{davis_marcus_2015}] are often used in AI programs.'' \cite[p.~95]{davis_marcus_2015}. ``Many specialized taxonomies have been developed in domains such as medicine and genomics.''({\em ibid}.~p.~96), and ``A number of sophisticated extensions of the basic inheritance architecture described here have also been developed. Perhaps the most powerful and widely used of these is description logic. Description logics provide tractable constructs for describing concepts and the relations between concepts, grounded in a well-defined logical formalism. They have been applied extensively in practice, most notably in the semantic Web ontology language OWL.'' ({\em ibid}.~p.~96).

As we saw in Section \ref{class-inclusion_and_part-whole_relations_section}, taxonomic relations may be expressed with SP-patterns, with pattern recognition at multiple levels of abstraction.

\begin{itemize}

    \item {\em Basic Relations}. With regard to the three basic taxonomic relations described by DM:

        \begin{itemize}

            \item {\em An individual is an instance of a category}. This can be seen in the way that the unknown plant in the example shown in Figure \ref{class_part_plant_figure} may be recognised as an instance of the species Meadow Buttercup, a member of the genus {\em Ranunculus}, a member of the family Ranunculaceae, and so on.

            \item {\em One category is a subset of another}. In the SP system, this kind of relationship is expressed using pairs of SP-symbols like `\texttt{<genus> ...~</genus>}' which, in the figure, provide the connection between the SP-pattern describing the concept `genus' (column 6 in the figure) and the SP-pattern describing the concept `species' (column 1).

            \item {\em Two categories are disjoint}. In the SP system, this kind of relationship would be implicit in most collections of SP-patterns. There appears to be no need for it to be marked explicitly.

        \end{itemize}

    \item {\em Categories and properties}. DM write that ``Categories can ...~be tagged with properties. For instance, \texttt{Mammal} is tagged as \texttt{Furry} [in their Figure 2].'' \cite[p.~95]{davis_marcus_2015}. In this connection, an important feature of the SP system is that there is no distinction between categories and properties (otherwise known as classes and attributes), and, in a similar way, there is no distinction between `classes' and `objects', a distinction which is prominent in object-oriented programming. In the SP system, all such concepts are modelled with `SP-patterns' and `SP-symbols'.\footnote{Although concepts like `category' and `property' may be used informally where appropriate.}

        There are three main advantages in removing these distinctions:

            \begin{itemize}

                \item It facilitates the seamless integration of class-inclusion relations with part-whole relations, as illustrated in Figure \ref{class_part_plant_figure}. Achieving that integration was one of the original motivations for the development of the SP system.\footnote{When I was employed in a software company and working on the development of a `support environment' for software engineers, it became clear that there was a need to integrate classes or versions of software products with the parts and sub-parts of such products, and that this was difficult to do with existing technologies.}

                \item If a class cannot be an object (which is a feature of some object-oriented systems), then there is a need for the category `metaclass' (a class of a class). This in turn points to the need for such categories as `metametaclass', `metametametaclass', and so on, in a rather unhelpful recursive loop \cite[Section 6.4.3.1]{wolff_2006}.

                \item Removing the distinctions facilitates the representation of cross-classification and other relatively complex kinds of taxonomy outlined in the third point under `Forms of inference', next.

            \end{itemize}

    \item {\em Forms of inference}. With regard to the taxonomic forms of inference discussed by DM:

        \begin{itemize}

            \item {\em Transitivity}. ``Since \texttt{Lassie} is an instance of \texttt{Dog} and \texttt{Dog} is a subset of \texttt{Mammal}, it follows that \texttt{Lassie} is an instance of \texttt{Mammal}.'' \cite[p.~95]{davis_marcus_2015}. In the SP system, relationships like those may be implicit in a set of SP-patterns describing different categories of animal. They would not be encoded explicitly.

            \item {\em Default inheritance}. ``A variant of [inheritance] is default inheritance; a category can be marked with a characteristic but not universal property, and a subcategory or instance will inherit the property unless it is specifically canceled.'' \cite[p.~95]{davis_marcus_2015}. As described in \cite[10.1]{sp_extended_overview} and \cite[Section 7.7]{wolff_2006}, the SP system can model the closely-related nonmonotonic reasoning with default values.

            \item {\em Other taxonomies}. ``Other taxonomies are less straightforward. For instance, in a semantic network for categories of people, the individual \texttt{GalileoGalilei} is simultaneously a \texttt{Physicist}, an \texttt{Astronomer}, a \texttt{ProfessorOfMathematics}, a \texttt{WriterInItalian}, a \texttt{NativeOfPisa}, a \texttt{PersonChargedWithHeresy}, and so on. These overlap, and {\em it is not clear which of these are best viewed as taxonomic categories and which are better viewed as properties}. In taxonomizing more abstract categories, choosing and delimiting categories becomes more problematic; for instance, in constructing a taxonomy for a theory of narrative, the membership, relations, and definitions of categories like \texttt{Event}, \texttt{Action}, \texttt{Process}, \texttt{Development}, and \texttt{Incident} are uncertain.'' (\cite[p.~95]{davis_marcus_2015}, emphasis added).

                The points that DM make here relate to two features of the SP system:

                \begin{itemize}

                    \item With SP-patterns within the SP-multiple-alignment framework, it is as straightforward to model cross-classification or class heterarchies as it is to model ordinary class hierarchies \cite[Section 6.4]{wolff_2006}.

                    \item That ``...~it is not clear which [of the attributes of Galileo] are best viewed as taxonomic categories and which are better viewed as properties'' (\cite[p.~95]{davis_marcus_2015}, emphasis in the original) lends support to the feature of the SP system (noted under `Categories and properties' in  Section \ref{taxonomic_reasoning_section}) that it avoids formal distinctions between such things as `categories' and `properties'

                \end{itemize}

        \end{itemize}

\end{itemize}

\subsubsection{Temporal Reasoning}\label{temporal_reasoning_section}

DM write:

\begin{quote}

    ``Representing knowledge and automating reasoning about times, durations, and time intervals is a largely solved problem. For instance, if one knows that Mozart was born earlier and died younger than Beethoven, one can infer that Mozart died earlier than Beethoven.~...~Integrating such reasoning with specific applications, such as natural language interpretation, has been ...~problematic.~...~many important temporal relations are not explicitly stated in texts, they are inferred; and the process of inference can be difficult.'' \cite[p.~96]{davis_marcus_2015}.

\end{quote}

No attempt has yet been made to apply the SP system to temporal reasoning, so the remarks that follow are tentative.

As described in \cite[Sections 6.1 and 6.2]{sp_vision}, the SP system has potential for the building of 3D digital models of objects and environments, and for spatial reasoning with such models. The suggestion here, partly motivated by what people seem to do in reasoning about temporal relations, is that such reasoning may be done in a manner that is similar to or the same as spatial reasoning---via the computational manipulation of digital objects representing periods of time. This does not solve the problem of interpreting natural language descriptions of a temporal reasoning problem, but it may throw some light on how temporal reasoning may be done after the natural language description has been translated into an appropriate form.

There is indirect support for this idea via the thinking behind Cuisenaire rods: wooden or plastic sticks in different colours and different lengths which are used as an aid in the teaching of arithmetic concepts to children---by showing how addition, subtraction, multiplication and division, and other arithmetic operations, may be understood visually.\footnote{See ``Cuisenaire rods'', {\em Wikipedia}, \href{http://bit.ly/2bOEQ6N}{bit.ly/2bOEQ6N}, retrieved 2016-09-05.} It seems possible that commonsense reasoning about temporal relations may be done, mentally or digitally, in a similar way.

\subsubsection{Action, Change, and Qualitative Reasoning}\label{action_change_qualitative_section}

DM describe two other areas of success in automated commonsense reasoning \cite[pp.~96--97]{davis_marcus_2015}:

\begin{itemize}

    \item {\em Action and change}. Modelling inferential processes related to actions, events, and change, with possibly over-simplified assumptions such as ``...~one event occurs at a time, and the reasoner need only consider the state of the world at the beginning and the end of the event, ...'', and ``Every change in the world is the result of an event.''

    \item {\em Qualitative reasoning}. Modelling forms of qualitative reasoning such as how the price of a product influences the number of items that are sold, and how an increase in the temperature of a gas in a closed container leads to an increase in pressure.

\end{itemize}

As with temporal reasoning, no attempt has yet been made to apply the SP system to these areas, so remarks about them are tentative---and they are reserved for the following subsection.

\subsubsection{Discussion}\label{taxonomic_reasoning_discussion_section}


As DM say, there has been some success in the four areas discussed above (Section \ref{successes_in_csr_section}). These successes may prove useful in future development of the SP system.

But there is one main shortcoming of the four areas of success in CSRK described by DM: they have apparently been developed quite independently of each other or other aspects of intelligence.

This is the kind of fragmentation in AI that Pamela McCorduck criticised so lucidly: ``The goals once articulated with debonair intellectual verve by AI pioneers appeared unreachable ...~Subfields broke off---vision, robotics, natural language processing, machine learning, decision theory---to pursue singular goals in solitary splendor, without reference to other kinds of intelligent behaviour.'' \cite[p.~417]{mccorduck_2004}. And the fragmentation is at odds with the central aim of the SP research, to promote simplification and integration of observations and concepts across a broad canvass (Appendix \ref{distinctive_features_and_advantages_appendix}).

The penalty of developing these four areas independently of each other is that there is little or no integration amongst them, and they appear to have little or nothing to say to each other. Since they all deal with aspects of CSRK, it is disappointing that there is no overarching conceptual framework or theory. And it would have been useful to have some insight into how the four areas might integrate with other aspects of intelligence, especially learning.

By contrast with the four areas of success described by DM, the SP system aims for integration across a much broader canvass and it appears to have potential in all the areas that DM have discussed, and for their integration.

An example that illustrates the potential benefits of large-scale integration is ``...~the problem of integrating action descriptions at different levels of abstraction.'' mentioned in the \cite{davis_marcus_2015} section about ``Action and change'' \cite[p.~96]{davis_marcus_2015}---although the examples given by DM seem to represent part-whole relations rather than levels of abstraction (class-inclusion relations). Either way, research on action, change, and qualitative reasoning, would probably benefit from ideas in the SP system about taxonomic reasoning and, more generally, how those constructs may be realised in the SP system (Section \ref{versatility_in_ai_section}).

\subsection{Challenges in Automating Commonsense Reasoning}


In a section with the heading above \cite[pp.~97--99]{davis_marcus_2015}, DM describe five challenges for the automation of commonsense reasoning, discussed in the following subsections.

\subsubsection{Many Domains Are Poorly Understood}\label{domains_poorly_understood_section}

\begin{quote}

    ``...~many of the domains involved in commonsense reasoning are only partially understood or virtually
    untouched.'' \cite[p.~97]{davis_marcus_2015}.

\end{quote}

This is true and \cite{davis_marcus_2015} does a valuable service in highlighting the complexities of CSRK and associated challenges for AI.

\subsubsection{Logical Complexity and The Horse's Head Scene in {\em The Godfather}}\label{logical_complexity_section}

DM's assertion that ``...~situations that seem straightforward can turn out, on examination, to have considerable logical complexity.'' \cite[p.~97]{davis_marcus_2015} is certainly true, and the example that they give---the horse's head scene in {\em The Godfather} that was mentioned in the Introduction---illustrates the point very well. Since this is an interesting and challenging example, this subsection expands on how the example may be analysed and how it may be modelled with the SP system.

In summary, the relevant part of the plot is this:

\begin{quote}

    ``Johnny Fontane, a famous singer and godson to Vito [Corleone---the Godfather], seeks Vito's help in securing a movie role; Vito dispatches his consigliere, Tom Hagen, to Los Angeles to talk the obnoxious studio head, Jack Woltz, into giving Johnny the part. Woltz refuses until he wakes up in bed with the severed head of his prized stallion.'' (Adapted from ``The Godfather'', {\em Wikipedia}, \href{http://bit.ly/2c5YZAy}{bit.ly/2c5YZAy}, retrieved 2016-09-12.)

\end{quote}

Instead of trying to understand the example from the perspective of a cinema audience, the analysis here will focus on how Jack Woltz might interpret the unpleasant experience of finding a horse's head in his bed.

Although recognition and inference are intimately related (Section \ref{inheritance_of_attributes_section}), it seems there would be those two main phases in Woltz's thinking:

\begin{itemize}

    \item {\em Phase 1: Recognition}.

    \begin{itemize}

        \item In order to make sense of the event, the first step is that Woltz must recognise the horse's head as what it is. This may seem too easy and simple to deserve comment but that should not disguise the existence of this first step or its complexity.

        \item The next step, which may again seem too simple to deserve comment, is that Woltz would make the very obvious inference that the horse's head had been part of a horse.

        \item Woltz would also recognise that the horse was his prized stallion which, we shall suppose, was called ``Lightning Force''. We shall suppose also that a white flash on the horse's forehead is distinctive for the stallion, although indirect inferences would probably also lead to the same identification.

    \end{itemize}

    \item {\em Phase 2: Inference}. Why should the head of Lightning Force have appeared in Woltz's bed? Here are some possibilities.

    \begin{itemize}

        \item It could have been some kind of accident, although it is much more likely that it was the deliberate act by some person.

        \item Assuming that it was a deliberate act, what was the motivation? Here, Woltz's knowledge of the Mafia would kick in: killing things is something that the Mafia do as a warning or means of persuading people to do what they want. The person to be persuaded must have an emotional attachment to the person or animal that is killed.\footnote{This is a little different from DM's interpretation: ``...~it is clear Tom Hagen is sending Jack Woltz a message---if I can decapitate your horse, I can decapitate you; cooperate, or else.'' \cite[p.~93]{davis_marcus_2015} but, arguably, equally valid.}

        \item Woltz also knows that Tom Hagen is a member of the Mafia and that Hagen wants Woltz to give Johnny Fontane a part in a movie. From that knowledge and his knowledge of how the Mafia operate, Woltz can make connections with the killing of Lightning Force.

    \end{itemize}

\end{itemize}

Figure \ref{horses_head_part1_figure} shows how Phase 1 in the scheme above (the recognition phase) may be modelled via the creation of an SP-multiple-alignment created by the SP Computer Model. In this example, the computer model has been supplied with a set of Old SP-patterns describing various aspects of horses, mammals, and of Lightning Force. It has also been supplied with New information describing some of the features of the severed head that Woltz saw.

\begin{figure}[!htbp]
\fontsize{07.00pt}{08.40pt}
\centering
{\bf
\begin{BVerbatim}
0             1                2                3              4             5

                                                                             <lf>
                                                                             lf1
                                                                             Lightning
                                                                             Force
                                                <horse> -------------------- <horse>
                                                h1
                               <mammal> ------- <mammal>
                               m1
                               <head> ------------------------ <head>
                                                               h2
                                                               <marks> ----- <marks>
white-flash ---------------------------------------------------------------- white-flash
                                                               </marks> ---- </marks>
long-snout --------------------------------------------------- long-snout
large-teeth -------------------------------------------------- large-teeth
                               </head> ----------------------- </head>
                               <body> --------- <body>
                                                hindgut
                                                fermentation
                               </body> -------- </body>
                               <legs> --------- <legs>
                                                odd-toed
                               </legs> -------- </legs>
              <vital-signs> -- <vital-signs>
              v2
dead -------- dead
              </vital-signs> - </vital-signs>
                               </mammal> ------ </mammal>
                                                </horse> ------------------- </horse>
                                                                             </lf>

0             1                2                3              4             5
\end{BVerbatim}
}
\caption{An SP-multiple-alignment, created by the SP Computer Model, for the recognition phase in the horse's head example, as discussed in the text.}
\label{horses_head_part1_figure}
\end{figure}

That set of features includes one that indicates that the horse is dead. Of course, this is a simplification of the way in which physical signs would have shown that the severed head, and thus the whole horse, was dead.

In the SP-multiple-alignment shown in Figure \ref{horses_head_part1_figure}, the New information, which describes what Woltz saw and which appears in column 0, makes connections with various parts of the Old SP-patterns in columns 1 to 5. The SP-multiple-alignment shows that the horse's head, represented by the SP-pattern in column 4, has been recognised, that it connects with the `\texttt{head}' part of an SP-pattern representing the structure of mammals (column 2), that this SP-pattern connects with an SP-pattern representing horses (column 3), and that this in turn connects with an SP-pattern representing Lightning Force (column 5). As mentioned above, we shall assume that Woltz recognises his prized stallion by the distinctive white flash on its forehead and it is this feature which brings the SP-pattern for Lightning Force into the SP-multiple-alignment (column 5).

Figure \ref{horses_head_part2_figure} shows an SP-multiple-alignment for Phase 2 in the scheme above (the inference phase), ignoring the possibility (the second point under Phase 2) that the horse's head in Woltz's bed was the result of some kind of accident.

In principle, there could be one SP-multiple-alignment for both the recognition and the inference phases but this would have been too big to show on one page. So it has been convenient to split the analysis into two SP-multiple-alignments corresponding to the posited two phases in Woltz's thinking.

\newgeometry{left=2cm,right=0cm}


\begin{figure}[!htbp]
\fontsize{05.00pt}{06.00pt}
\centering
{\bf
\begin{BVerbatim}
0           1           2       3            4        5        6           7            8          9       10          11       12       13

                                                                           <psn>
                                                                           psn3 ------- psn3
                                                                           Tom
                                                                           Hagen
                                <mafiosi> -------------------------------- <mafiosi>
                                mfi1
                                Mafiosi
                                if
                                <x> ---------------------------------------------------------------------------------- <x>
                                                                                                                       x1
                                                                                                                       <psn> -- <psn>
                                                                                                                                psn2 --- psn2
                                                                                                                                Jack
                                                                                                                                Woltz
                                                                                                                       </psn> - </psn>
                                </x> --------------------------------------------------------------------------------- </x>
                                loves -------------------------------------------------------------------------------------------------- loves
                                <z> -------------------------------------------------------------- <z>
                                                                                                   z1
                                                                                                   <lf> -- <lf>
                                                                                                   lf1 --- lf1 ------------------------- lf1
                                                                                                           Lightning
                                                                                                           Force
                                                                                                           dead
                                                                                                   </lf> - </lf>
                                </z> ------------------------------------------------------------- </z>
                                persuade ---------------------------------------------- persuade
                                <x> -------- <x>
                                             x1
                                             <psn> -- <psn>
                                                      psn2 ---------------------------- psn2
                                                      Jack
                                                      Woltz
                                             </psn> - </psn>
                                </x> ------- </x>
                                to
                                do
                                <action> --------------------- <action>
                                                               ac2 -------------------- ac2
                                                               Give
                                                               Johnny
                                                               the
                                                               part
                                </action> -------------------- </action>
                                by
                                making
                        <z> --- <z>
                        z1
            <lf> ------ <lf>
            lf1 ------- lf1
Lightning - Lightning
Force ----- Force
dead ------ dead -------------- dead
            </lf> ----- </lf>
                        </z> -- </z>
                                </mafiosi> ------------------------------- </mafiosi>
                                                                           </psn>

0           1           2       3            4        5        6           7            8          9       10          11       12       13
\end{BVerbatim}
}
\caption{An SP-multiple-alignment for inference in the horse's head example, as discussed in the text.}
\label{horses_head_part2_figure}
\end{figure}


\restoregeometry

Probably the most important feature of the SP-multiple-alignment shown in Figure \ref{horses_head_part2_figure} is the SP-pattern shown in column 3 which describes a supposed feature of how the Mafiosi operate. This is, reading from the top, that if $x$ loves $z$, a member of the Mafiosi may persuade $x$ to do something (an `\texttt{action}' in the SP-pattern) by killing $z$ (the thing that $x$ loves). This is, no doubt, a distortion and oversimplification of how the Mafiosi operate but it is perhaps good enough for present purposes.

Other features of this SP-multiple-alignment concept include:

\begin{itemize}

    \item The SP-pattern for Tom Hagen (in column 7) connects with the SP-pattern for Mafiosi (column 3) and thus inherits their modes of operation.

    \item The SP-pattern in column 13 shows that Jack Woltz (with the reference code `\texttt{psn2}' in the SP-pattern for Jack Woltz in column 12) `\texttt{loves}' Lightning Force (with the reference code `\texttt{lf1}' in the SP-pattern for Lightning Force in column 10).

    \item That fact (that Jack Woltz loves Lightning Force) connects with ``$x$ loves $z$'' in the SP-pattern in column 3.

    \item Reading from the top, the SP-pattern in column 8 records the fact that Tom Hagen (with the reference code `\texttt{psn3}') is seeking to `\texttt{persuade}' Jack Woltz (with the reference code `\texttt{psn2}') to perform a particular `\texttt{action}' (with the reference code `\texttt{ac2}'). That action is to ``Give Johnny the part''.

\end{itemize}

The analysis of the horse's head scene that has been presented in this section is certainly not the last word, but I believe it suggests a possible way forward. Ultimately, robust capabilities will be needed for the unsupervised learning of CSK in realistic settings so that CSR may operate with relatively large and well-structured bodies of knowledge.

\subsubsection{Plausible Reasoning}\label{plausible_reasoning_section}

\begin{quote}

    ``...~commonsense reasoning almost always involves plausible reasoning; that is, coming to conclusions that are reasonable given what is known, but not guaranteed to be correct. Plausible reasoning has been extensively studied for many years, and many theories have been developed, including probabilistic reasoning, belief revision, and default reasoning or non-monotonic logic. However, overall we do not seem to be very close to a comprehensive solution. Plausible reasoning takes many different forms, including using unreliable data; using rules whose conclusions are likely but not certain; default assumptions; assuming one's information is complete; reasoning from missing information; reasoning from similar cases; reasoning from typical cases; and others. How to do all these forms of reasoning [perform] acceptably well in all commonsense situations and how to integrate these different kinds of reasoning are very much unsolved problems.'' \cite[p.~98]{davis_marcus_2015}.

\end{quote}

In the following two subsections, I argue, first, that the SP system shows promise as a means of modelling the kinds of reasoning mentioned in the quotation (and others), and, second, that it promises to solve the problem of integration, mentioned at the end of the quotation.

\begin{itemize}

    \item {\em Modelling different kinds of reasoning}. Some of the kinds of reasoning mentioned in the quotation above as being actual or potential elements of CSR seem to be the same as or similar to kinds of reasoning that have been demonstrate with the SP system (Section \ref{versatility_in_reasoning_section}). There are reasons to believe that most of the kinds of reasoning mentioned by DM may be accommodated by the SP system:

        \begin{itemize}

            \item {\em Probabilistic reasoning}. As noted in Appendix \ref{outline_of_sp_system_appendix}, the SP system is fundamentally probabilistic, so all kinds of reasoning in the SP system are probabilistic---although the system can, if required, imitate the clockwork nature of logical reasoning ({\em ibid.}).

            \item {\em Belief revision}. Since the SP system has strengths in the modelling of nonmonotonic reasoning (\cite[Section 10.1]{sp_extended_overview}, \cite[Section 7.7]{wolff_2006}), and since nonmonotonic reasoning has elements of belief revision (learning that Tweety is a penguin leads us to revise our earlier belief that Tweety can fly), there are reasons to believe that the SP system may serve to model other aspects of belief revision.

            \item {\em Default reasoning or non-monotonic logic}. As just noted, the SP system, with appropriate SP-patterns, may model nonmonotonic reasoning.

            \item {\em Reasoning using unreliable data}. A general feature of the SP system is that it can deliver plausible results with New information containing errors of omission, commission, and substitution (Section \ref{recognition_despite_errors_section}). Because of the generality of this feature in the SP system, there is reason to believe that it will also apply in reasoning.

            \item {\em Reasoning using rules whose conclusions are likely but not certain}. Because of the probabilistic nature of the SP system (Appendix \ref{origins_development_appendix}), most of the rules used in reasoning with the system would, normally, have conclusions that are likely but not certain.

            \item {\em Reasoning with default assumptions}. As noted above, the capabilities of the SP system with nonmonotonic reasoning demonstrate how it can reason with default assumptions, such as the assumption that, without contrary evidence, we may assume that if Tweety is a bird then he or she can fly.

            \item {\em Reasoning assuming one's information is complete}. If, in relevant databases, a travel agent cannot find a direct flight between two cities, then he or she would normally tell the customer that such a flight does not exist. In a similar way, the SP system would normally be used with the ``negation as failure'' assumption that, if information cannot be found within the system, then that information does not exist.\footnote{See ``Negation as failure'', {\em Wikipedia}, \href{http://bit.ly/2c0Ni36}{bit.ly/2c0Ni36}, retrieved 2016-09-10.}

            \item {\em Reasoning from missing information}. This aspect of reasoning has not yet been explored in the SP programme of research. The SP theory and the SP Computer Model probably need to be augmented to accommodate the notion of ``missing information''---because that notion provides a means of encoding information economically, in keeping with the principles on which the SP system is founded, but not yet incorporated in the SP system.

                To see why the concept of missing information provides a means of encoding information economically,
                consider how one would record the names of 10 people, all of whom are in the village football team.
                One could of course list them individually. But it would be more economical to record the list as
                something like ``Bloomfield Rovers, without Jack'', where Jack is the member of
                the football team who is not on the list.

            \item {\em Reasoning from similar cases, and reasoning from typical cases}. Where there are similarities amongst a set of cases, or where one or more cases can be recognised as being ``typical'' on the strength of similarities across the range of cases, then unsupervised learning in the SP system would identify redundancies amongst the several cases and, via lossless information compression, create an abstract representation or ``grammar'' for those cases. That grammar would provide the basis for reasoning with those cases and, since the grammar would normally generalise beyond the cases it was derived from \cite{wolff_1988}, it would normally provide the basis for reasoning with other cases that may be described by the grammar.

        \end{itemize}

        Further evidence that the SP system has potential as a vehicle for CSR may be seen in its strengths and potential in
        other kinds of reasoning which have the flavour of CSR, not mentioned in the quotation above but amongst those
        mentioned in Section \ref{versatility_in_reasoning_section}:

            \begin{itemize}

                \item {\em Bayesian reasoning with `explaining away'}.

                \item {\em Causal reasoning}.

                \item {\em Reasoning that is not supported by evidence}.

                \item {\em Inference via inheritance of attributes}.

                \item {\em Spatial reasoning}.

                \item {\em What-if reasoning}.

            \end{itemize}

        \item {\em Seamless Integration of {CSR}}. As described in Section \ref{seamless_integration_section} and elsewhere in this paper, the use of one simple format for the representation of all kinds of knowledge and one relatively simple framework---SP-multiple-alignment---for the processing of knowledge, are likely to facilitate the seamless integration of diverse kinds of knowledge and diverse aspects of intelligence.

            These remarks apply with equal force to the several forms of reasoning within the actual or potential capabilities of
            the SP system (Sections \ref{versatility_in_reasoning_section} and `Modelling Different Kinds of Reasoning' under
            Section \ref{plausible_reasoning_section}).

            As noted in Section \ref{seamless_integration_section}, ordinary experience suggests that seamless integration of
            diverse kinds of knowledge and diverse forms of reasoning are pre-requisites for the kinds of commonsense reasoning
            which we do constantly in everyday situations.

\end{itemize}

\subsubsection{Long Tail}\label{long_tail_section}

\begin{quote}

    ``...~in many domains, a small number of examples are highly frequent, while there is a `long tail' of a vast number of highly infrequent examples ... it is often very difficult to attain high quality results [in CSRK], because a significant fraction of the problems that arise correspond to very infrequent categories..'' \cite[p.~98]{davis_marcus_2015}.

\end{quote}

Probably the most famous example of the ``long tail'' phenomenon is the sentence {\em Colorless green ideas sleep furiously} that Noam Chomsky \cite[p.~15]{chomsky_1957} presented to illustrate, {\em inter alia}, how a sentence that is grammatical can be vanishingly rare. It is true that the long tail phenomenon is normally discussed in terms of relatively small entities such as words. But we may expand our view from words to sentences, as DM appear to recognise when they say ``In natural language text, for example, some trigrams (for example, `of the year') are very frequent, but many other possible trigrams, such as `moldy blueberry soda' or `gymnasts writing novels' are immediately understandable, yet vanishingly rare.'' \cite[p.~99]{davis_marcus_2015}.

When we view the frequencies of occurrence of different sentences as examples of the long tail phenomenon, we find that the great majority of sentences in any corpus occur only once in that corpus. Indeed, it is widely accepted that most sentence in most natural languages are new to the world.\footnote{In brief, reasons in support of that conclusion are as follows: since recursive structures are prominent in most natural languages, and since such structures can produce infinitely many surface forms, there is an infinite number of different possible sentences in any natural language. This is larger, by a wide margin, than the finite albeit large number of sentences that have actually been written or spoken.}

On the strength of examples like ``Colorless green ideas ...'', and the extraordinary complexity of natural language, Chomsky \cite{chomsky_1965}, followed by others, developed the `nativist' view that much of the structure of natural language is inborn. However, models of the unsupervised learning of language via the compression of information demonstrate how it is possible to learn a generative grammar via information compression despite the fact that some sentences that fall within the scope of the given grammar are vanishingly rare in the data from which the grammar was derived \cite{wolff_1982,wolff_1988}.

On the strength of evidence like that, it appears that Chomsky's assertion that ``...~one's ability to produce and recognize grammatical utterances is not based on notions of statistical approximation and the like.'' \cite[p.~16]{chomsky_1957} is wrong. It appears also that there may be no need to worry about ``a `long tail' of a vast number of highly infrequent examples.'' \cite[p.~98]{davis_marcus_2015}.

Another way of looking at these issues is via the concept of generalisation. Unsupervised learning via information compression, which includes unsupervised learning in the SP system, provides a persuasive account of how it is possible, without correction by a `teacher' or equivalent assistance, to create a grammar that generalises beyond the raw data without over-generalising (\cite[pp.~181--191]{wolff_1988}, \cite[Chapter 9]{wolff_2006}, \cite[Section 5.3]{sp_extended_overview}). In brief, the products of such learning are: a {\em grammar} (which provides an abstract description of the raw data); and an {\em encoding} of the raw data in terms of the grammar. The two things together provide lossless compression of the raw data. And, normally, the grammar generalises beyond the raw data without overgeneralising.

The significance of these observations for the long-tail phenomenon and CSR, which applies to any kind of data, not just natural language, is that generalisations beyond the raw data, of which there may be many, are  likely to be very rare in most samples of raw data, or entirely absent from all but the very largest samples of such data.

But it appears that the rarity of many generalisations is of little or no consequence for CSR, provided that the grammar captures recurring features of the world, as would normally be the case with unsupervised learning via information compression.

Consider, for example, the commonsense argument that if \texttt{A} comes before \texttt{B}, and \texttt{B} comes before \texttt{C}, then \texttt{A} comes before \texttt{C}. This kind of relationship is a recurring feature of the world, regardless of the rarity of specific examples of \texttt{A}, \texttt{B}, and \texttt{C}, individually or in combination. Those examples might, for example, be the very rare combination \texttt{elephant}, \texttt{screwdriver}, and \texttt{veggie burger} but the rarity of such a combination does not disturb the general rule (in any sequence `\texttt{A B C}', `\texttt{A}' always precedes `\texttt{C}') of which it is an example, and it is that rule and how it may be learned which is important for CSR.

\subsubsection{Discerning the Proper Level of Abstraction}\label{proper_level_of_abstraction}

\begin{quote}

    ``...~in formulating knowledge it is often difficult to discern the proper level of abstraction. Recall the example of sticking a pin into a carrot and the task of reasoning that this action may well create a hole in the carrot but not create a hole in the pin.~...~The question is, how broadly should such rules should be formulated?'' \cite[pp.~98--99]{davis_marcus_2015}.

\end{quote}

In brief, the putative ``SP'' answer to the ``proper level of abstraction'' and ``how broadly should ...~rules ...~be formulated'' is compression of information. In keeping with the argument in Section \ref{taxonomic_reasoning_discussion_section}---that we should not attempt to determine knowledge structures via analysis but should be guided by what emerges from a well-constructed learning system founded on ICSPMA---we should see what levels of abstraction emerge from learning via ICSPMA. By hypothesis, these would represent the proper levels of abstraction, where criteria for ``proper'' would include succinctness and naturalness in CSK (in accordance with the DONSVIC principle) and effectiveness and efficiency in CSR.

\subsubsection{Methodological and Sociological Obstacles}\label{methodological_sociological_section}

\begin{quote}

    ``A final reason for the slow progress in automating commonsense knowledge is both methodological and sociological. Piecemeal commonsense knowledge (for example, specific facts) is relatively easy to acquire, but often of little use, because of the long-tail phenomenon discussed previously. Consequently, there may not be much value in being able to do a little commonsense reasoning.'' \cite[p.~99]{davis_marcus_2015}.

\end{quote}

The main points that DM make here are puzzling. It's not clear why piecemeal commonsense knowledge should be of little use or why the long-tail phenomenon is relevant. A young child is likely to learn quickly that getting burned is painful and that food is normally nice, and such knowledge is likely to prove useful despite the fact that they are specific facts that are largely independent of the many different ways of getting burned and the many different things that are edible. And they connect with frequently-occurring situations that are not in the ``long-tail'' category.

\subsection{Objectives For Research in {CSRK}}\label{objectives_section}

This section contains some brief comments on how the SP system relates to DM's objectives for research in commonsense reasoning \cite[pp.~99--100]{davis_marcus_2015}:

\begin{itemize}

    \item {\em Reasoning architecture}. This means ``The development of general-purpose data structures for encoding knowledge and algorithms and techniques for carrying out reasoning.'' \cite[p.~99]{davis_marcus_2015}.

        In the SP system, the SP-multiple-alignment framework with SP-patterns has proved to be a versatile system for the representation
        of knowledge (Section \ref{versatility_in_rk_section}), for several aspects of intelligence excluding reasoning (Section \ref{versatility_in_ai_section}), including several forms of reasoning (Section \ref{versatility_in_reasoning_section}).

    \item {\em Plausible inference}. ``Drawing provisional or uncertain conclusions'' \cite[p.~99]{davis_marcus_2015} is central in the workings of the SP system since the system is fundamentally probabilistic (Appendix \ref{outline_of_sp_system_appendix}).

    \item {\em Range of reasoning modes}. With regard to this objective---``incorporating a variety of different modes of inference, such as explanation, generalization, abstraction, analogy, and simulation'' \cite[p.~99]{davis_marcus_2015}:

        \begin{itemize}

            \item  ``Explanation'' is an implicit part of unsupervised learning in the SP system since a main product of that learning is a ``grammar'' which may be regarded as theory of raw data from which interpretations or explanations may be drawn via the building of SP-multiple-alignments.

            \item ``Generalisation'' is an important part of unsupervised learning by the SP system as outlined in Section \ref{long_tail_section}.

            \item ``Abstraction'' is a fundamental part of unsupervised learning by the SP system.

            \item ``Analogy'' has not been addressed directly in the SP programme of research, but the SP system is clearly relevant to this topic because of its ability to recognise similarities between patterns, outlined in Sections \ref{class-inclusion_and_part-whole_relations_section} and \ref{recognition_despite_errors_section}.

            \item Again, ``simulation'' has not been an explicit focus of interest in the SP programme to date, but the system is relevant to that topic because a grammar that has been abstracted from a given body of raw data via unsupervised learning provides a means of simulating the source or sources of those data.

        \end{itemize}

    \item {\em Painstaking analysis of fundamental domains}. ``In doing commonsense reasoning, people are able to do complex reasoning about basic domains such as time, space, na{\"i}ve physics, and na{\"i}ve psychology. The knowledge they are drawing on is largely unverbalized and the reasoning processes largely unavailable to introspection. An automated reasoner will have to have comparable abilities.'' \cite[pp.~99]{davis_marcus_2015}.

        Clearly, careful analysis of human CSRK will be needed for the successful automation of such knowledge representation and reasoning in
        artificial systems, including the SP system.

    \item {\em Breadth}. ``Attaining powerful commonsense reasoning will require a large body of knowledge.'' (({\em ibid}). This is clearly true for all artificial systems including the SP system.

    \item {\em Independence of experts}. ``Paying experts to hand-code a large knowledge base is slow and expensive. Assembling the knowledge base either automatically or by drawing on the knowledge of non-experts is much more efficient.'' (({\em ibid}).

        The strengths and potential of the SP system in unsupervised learning is likely to prove useful in the automatic learning of knowledge.
        Learning from books and other written material is likely to be important for CSRK and here the strength and potential of the SP system
        in the interpretation of natural language is clearly relevant, although substantial work will be needed to develop true understanding of
        text, not the relatively superficial processing in IBM's Watson, mentioned by DM \cite[p.~94]{davis_marcus_2015} (see also \cite[Section
        IX]{sp_alternatives}).

    \item {\em Applications}. ``To be useful, the commonsense reasoner must serve the needs of applications and must interface with them smoothly.'' \cite[p.~99]{davis_marcus_2015}.

        Since it is envisaged that, in mature versions of the SP system, all applications and CSRK will be realised via SP-patterns in the
        SP-multiple-alignment framework, and since the one simple format for knowledge and the one relatively simple framework for the
        processing of knowledge is likely to facilitate the seamless integration of knowledge and processing (Section
        \ref{seamless_integration_section}), there are reasons to believe that the SP system will facilitate the smooth interfacing of the
        commonsense reasoner with diverse applications, all of them hosted on the SP system.

    \item {\em Cognitive modelling}. The SP programme of research is founded on earlier research that highlights the significance of information compression in the workings of brains and nervous systems and in children's learning of natural language (Appendix \ref{outline_of_sp_system_appendix}).

\end{itemize}

\section{Conclusion}

Understanding commonsense reasoning (CSR) and commonsense knowledge (CSK) is indeed challenging, but the {\em SP Theory of Intelligence} and its realisation in the {\em SP Computer Model} have relevant strengths and potential.

In brief, the main elements of this paper are:

\begin{itemize}

    \item {\em Other research on commonsense reasoning and commonsense knowledge}. This main section briefly reviews other research on commonsense reasoning and commonsense knowledge.

    \item {\em Why the SP system may prove useful with commonsense reasoning and commonsense knowledge}. This main section describes features of the SP system that suggest that it has strengths and potential in CSR and CSK:

        \begin{itemize}

            \item {\em Simplicity and Power}. The SP system is the product of a unique attempt to simplify and integrate observations and concepts across a wide area (Appendix \ref{distinctive_features_and_advantages_appendix}). The SP system, with the powerful concept of {\em SP-multiple-alignment} at centre stage, provides a much more favourable combination of conceptual {\em Simplicity} with descriptive or explanatory {\em Power} than any alternatives, including the kinds of CSRK-related systems outlined in the previous main section. To the extent that the SP system provides solutions to problems in CSRK, it is likely that those solutions will integrate smoothly amongst themselves and with other kinds of cognitive processing, a kind of seamless integration that appears to be essential in any theory of CSRK, or indeed AI.

            \item {\em Turing equivalence plus aspects of human intelligence}. It appears that the SP system has the generality of a universal Turing machine but with the kinds of strengths in human-like intelligence that Alan Turing recognised were missing from the universal Turing machine. This is a kind of generality that appears to be needed in CSR.

            \item {\em Information compression}. Information compression, which is central in the workings of the SP system, is likely to facilitate CSR and CSK because:
                \begin{itemize}

                    \item The intimate relation that is known to exist between information compression and concepts of inference and probability (Appendix \ref{origins_development_appendix}) is in keeping with the probabilistic nature of CSR.

                    \item The generality of the SP-multiple-alignment concept may provide the generality needed for CSRK.

                    \item The DONSVIC principle \cite[Section 5.2]{sp_extended_overview} suggests that knowledge learned by the SP system is likely to have the {\em succinctness} and {\em naturalness} that may be expected in CSK.

                \end{itemize}

            \item {\em Versatility in aspects of intelligence}. The SP system has strengths and potential in several aspects of intelligence including: unsupervised learning; natural language processing; fuzzy pattern recognition; recognition at multiple levels of abstraction; best-match and semantic forms of information retrieval; several kinds of reasoning (more below); planning; and problem solving.

                Examples from the SP Computer Model presented in this section show how the SP system may model pattern recognition, class-inclusion relations and part-whole relations, inheritance of attributes, and recognition in the face of errors of omission, commission, or substitution.

            \item {\em Versatility in kinds of reasoning}. Strengths and potential of the SP system in reasoning include: one-step `deductive' reasoning; chains of reasoning; abductive reasoning; reasoning with probabilistic networks and trees; reasoning with `rules'; nonmonotonic reasoning and reasoning with default values; Bayesian reasoning with `explaining away'; causal reasoning; reasoning that is not supported by evidence; inheritance of attributes; spatial reasoning; and what-if reasoning.

            \item {\em Versatility in the representation of knowledge}. Within the SP-multiple-alignment framework, SP-patterns have proved to be effective in representing several forms of knowledge, any of which may serve in CSK: the syntax of natural language; class hierarchies, class heterarchies (meaning class hierarchies with cross classification); part-whole hierarchies; discrimination networks and trees; entity-relationship structures; relational knowledge; rules for use in reasoning; SP-patterns in one or two dimensions; images; structures in three dimensions; and procedural knowledge.

            \item {\em Seamless integration of diverse aspects of intelligence and diverse forms of knowledge in any combination}. The use of one simple format for knowledge and one relatively simple framework for the processing of knowledge promotes seamless integration of diverse aspects of intelligence and diverse forms of knowledge, in any combination. That kind of integration appears to be essential for CSRK.

        \end{itemize}

    \item {\em Aspects of commonsense reasoning and commonsense knowledge and how they may be modelled in the SP system}. The discussion in this section focuses mainly on what DM say about CSR and CSK and how the SP system may meet those challenges:

        \begin{itemize}

            \item {\em Father and son, and other examples}. This section discusses how, via the `inheritance of attributes' mode of reasoning, the SP system may model several of DM's examples of CSR, including the way people instantly know the answer to questions like ``Can you make a salad out of a polyester shirt?''.

            \item {\em Commonsense in intelligent tasks}. Here, the SP system's strengths with `inheritance of attributes' may provide part of the explanation of how people disambiguate pronouns in the `Winograd Schema' and elsewhere. Other strengths of the SP system in CSR-related tasks include the way it can model the role of context in recognition and our ability to `see' things which are not objectively present in a picture. The system's strengths in diverse forms of reasoning may prove useful in the control of robots.

            \item {\em Successes in automated commonsense reasoning}. Also discussed are current successes in the automation of CSR (taxonomic reasoning, temporal reasoning, action and change, and qualitative reasoning), how the SP system's strengths in simplification and integration may promote seamless integration across these areas, and seamless integration of those area with other aspects of intelligence.

            \item {\em Challenges in automating commonsense reasoning}. The paper considers how the SP system may help overcome some of the challenges in the automation of CSR described by DM including: the logical complexity of much of CSR, including the `horse's head' scene in the film {\em The Godfather}---which is discussed at length with an example of how the SP system may model processes of interpretation and reasoning associated with this scene; modelling aspects of `plausible reasoning'; and modelling the learning required for CSR, with a solution of the `long tail' problem.

            \item {\em Objectives for research in CSR and CSK}. The SP system has what appear to be useful things to say about several of DM's objectives for research in CSR and CSK: the development of a general-purpose reasoning architecture; how to draw provisional or uncertain conclusions; how to incorporate a variety of different modes of inference; how reasoning may integrate smoothly with applications; and the need for consistency with human cognition. Unsupervised learning by the SP system may largely automate the process of obtaining the knowledge needed for the modelling of CSR and CSK.

        \end{itemize}

\end{itemize}

\section*{Acknowledgements}

I'm grateful for constructive comments by anonymous referees on earlier drafts of this paper.

\appendix

\section{Outline of the {SP} System}\label{outline_of_sp_system_appendix}

This Appendix provides an outline description of the SP system. More information, listed here in increasing levels of detail, may be found in \cite{sp_intro_2018}, \cite{sp_extended_overview}, and \cite{wolff_2006}. Other papers in the SP programme of research, including several about potential benefits and applications of the system, are detailed with download links near the top of \href{http://www.cognitionresearch.org/sp.htm}{www.cognitionresearch.org/sp.htm}.

\subsection{Backround: Origins and Development of the SP System}\label{origins_development_appendix}

The SP programme of research has been inspired in part by an earlier programme of research developing computer models of language learning, summarised in \cite{wolff_1988}. A key idea in that earlier research was learning via the identification of recurrent `chunks' of information \cite{miller_1956}, including the identification of chunks containing other chunks, leading to the creation of hierarchical (tree-structured) kinds of procedural knowledge.

With the new goal of the SP research---simplification and integration of observations and concepts across a broad canvass (Appendix \ref{distinctive_features_and_advantages_appendix})---hierarchical chunking would not do. The aim has been to discover or create a conceptual framework that would accommodate a wide variety of aspects of intelligence and a wide variety of kinds of knowledge, including both tree-structured and non-tree-structured kinds of knowledge, both procedural and static. As outlined in Appendix \ref{distinctive_features_and_advantages_appendix}, this quest has been largely successful, with the discovery and development of the powerful concept of {\em SP-multiple-alignment}, borrowed and adapted from the concept of `multiple sequence alignment' in bioinformatics.

The SP research, like the earlier research on language learning, has been inspired in part by a body of research, pioneered by Fred Attneave \cite{attneave_1954}, Horace Barlow \cite{barlow_1959,barlow_1969}, and others, with a focus on the importance of information compression in human learning, perception, and cognition. More information in this area may be found in \cite{sp_compression}.

Another significant strand of thinking in the development of the SP system is the intimate relation that is known to exist between information compression and concepts of inference and probability \cite{shannon_1948,solomonoff_1964,solomonoff_1997,li_vitanyi_2014}. This close relation means that, although the SP system is dedicated to information compression, it has significant strengths in the making of inferences (\cite[Section Section 10]{sp_extended_overview}, \cite[Chapter 7]{wolff_2006}) and in the calculation of probabilities (\cite[Section 4.4]{sp_extended_overview}, \cite[Section 3.7]{wolff_2006}).

Because the SP system, in its operation, is dedicated to information compression, the name `SP' is short for {\em Simplicity} and {\em Power}. This is because information compression may be seen to be a process of maximising the {\em simplicity} of a body of information, {\bf I}, by extracting redundancy from {\bf I}, whilst retaining as much as possible of its non-redundant descriptive {\em power}.

At present, the SP Theory of Intelligence is realised in the form of the SP Computer Model. The Theory and the Model have been developed together, with the computer model helping to reduce vagueness in the theory, and providing a means of testing the theory and demonstrating what it can do. Testing the theory in the form of a computer model has proved to be essential as a means of weeding out bad ideas. Many such ideas have been discarded as a result of testing in a long succession of versions of the SP Computer Model over a period of about 17 years. The current SP Computer Model provides a relatively robust and well-validated expression of the SP Theory. Source code and Windows executive code for the SP Computer Model may be downloaded from below the heading ``SOURCE CODE'' on \href{http://www.cognitionresearch.org/sp.htm}{www.cognitionresearch.org/sp.htm}.

\begin{figure}[!htbp]
\centering
\includegraphics[width=0.9\textwidth]{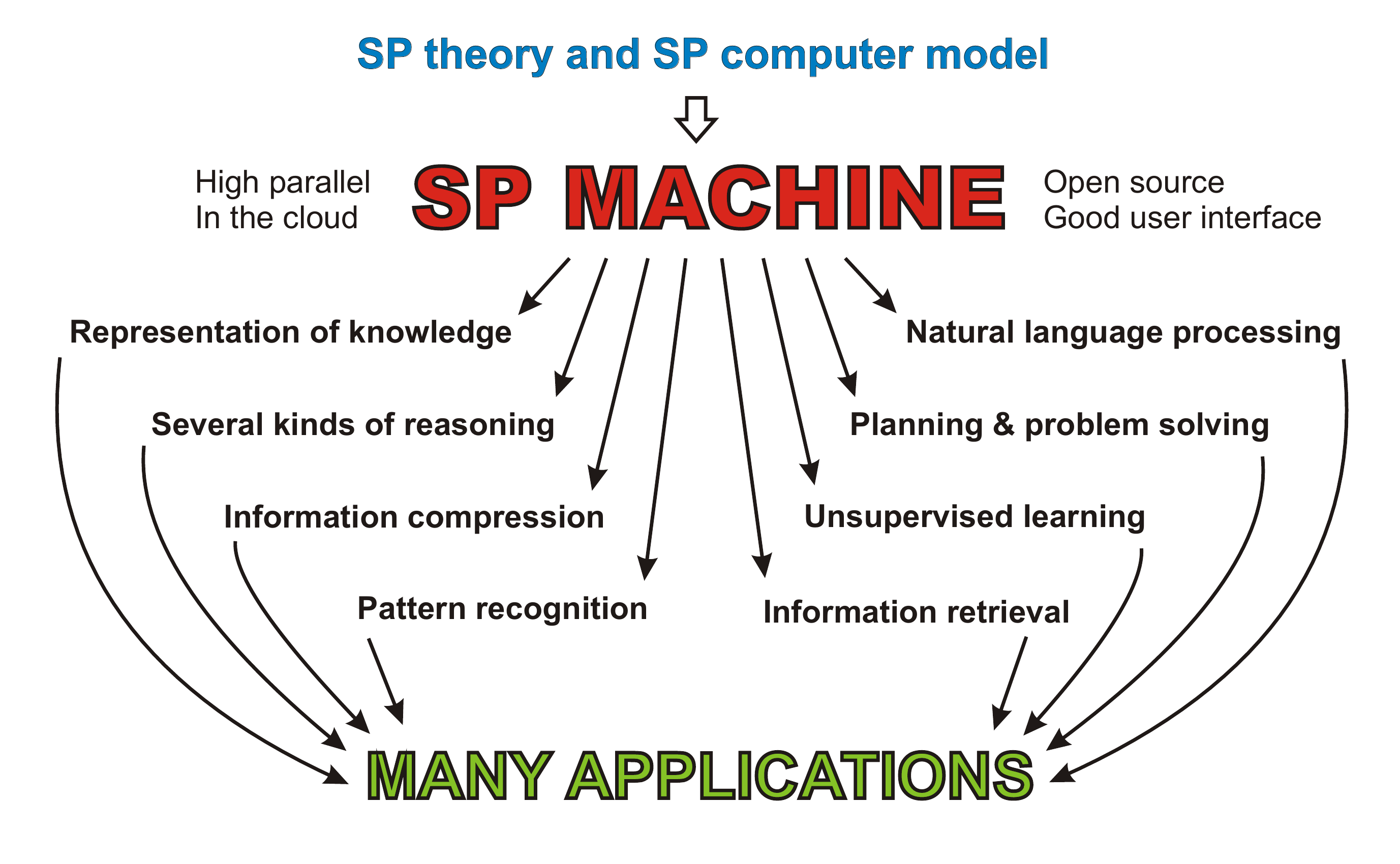}
\caption{Schematic representation of the development and application of the SP machine. Reproduced from Figure 2 in \cite{sp_extended_overview}, with permission.}
\label{sp_machine_figure}
\end{figure}

It is envisaged that the SP Theory and the SP Computer Model will provide the basis for an industrial-strength {\em SP Machine}, as shown schematically in Figure \ref{sp_machine_figure}.

Things to be done in the development of such an SP machine are described in \cite{spagi_2017}.

\subsection{Distinctive Features and Potential Advantages of the SP System}\label{distinctive_features_and_advantages_appendix}

{\em The {\bf SP theory of Intelligence} and its realisation in the {\bf SP Computer Model} is the product of a unique programme of research, seeking to simplify and integrate observations and concepts across artificial intelligence, mainstream computing, mathematics, and human learning, perception, and cognition.} The overall aim has been to discover or invent a system that combines {\em Simplicity} with relatively high descriptive or explanatory {\em Power}.

As mentioned above, a key development from this research is the remarkably simple but powerful concept of {\em SP-multiple-alignment}, which is the foundation for most of the strengths and potential in the SP system. All the examples of SP-multiple-alignments in this paper are output from the SP Computer Model.

In brief, the SP system has strengths and potential in:

\begin{itemize}

    \item {\em Several aspects of intelligence} including: unsupervised learning; the  analysis and  production  of  natural  language;  pattern  recognition and more summarised in Section \ref{versatility_in_ai_section}.
    \item {\em Several kinds of reasoning} including: one-step  `deductive'  reasoning; chains of reasoning; abductive reasoning; reasoning with probabilistic networks and trees; and more, summarised in Section \ref{versatility_in_reasoning_section}.
    \item {\em The representation of several different kinds of knowledge}, including: the syntax of natural languages; class-inclusion hierarchies (with or without cross classification); part-whole hierarchies; and more, summarised in Section \ref{versatility_in_rk_section}.
    \item Because the SP system's strengths and potential in diverse aspects of intelligence (including diverse kinds of reasoning), and in the representation diverse forms of knowledge, all flow from one relatively simple framework---SP-multiple-alignment---{\em there is potential for the seamless integration of diverse aspects of intelligence and diverse kinds of knowledge, in any combination} as described in Section \ref{seamless_integration_section}.
    \item {\em Potential benefits and applications of the SP system, and other strengths of the system} are summarised in \cite[Sections 7 to 9]{sp_intro_2018}.
    \item {SP-Neural}. Abstract concepts in the SP theory may be mapped into a version of the theory called {\em SP-Neural} expressed in terms of neurons and their inter-connections. This is described relatively briefly in \cite[Chapter 11]{wolff_2006} and with much more detail in \cite{spneural_2016}.

\end{itemize}

\begin{figure}[!hbt]
\centering
\includegraphics[width=0.9\textwidth]{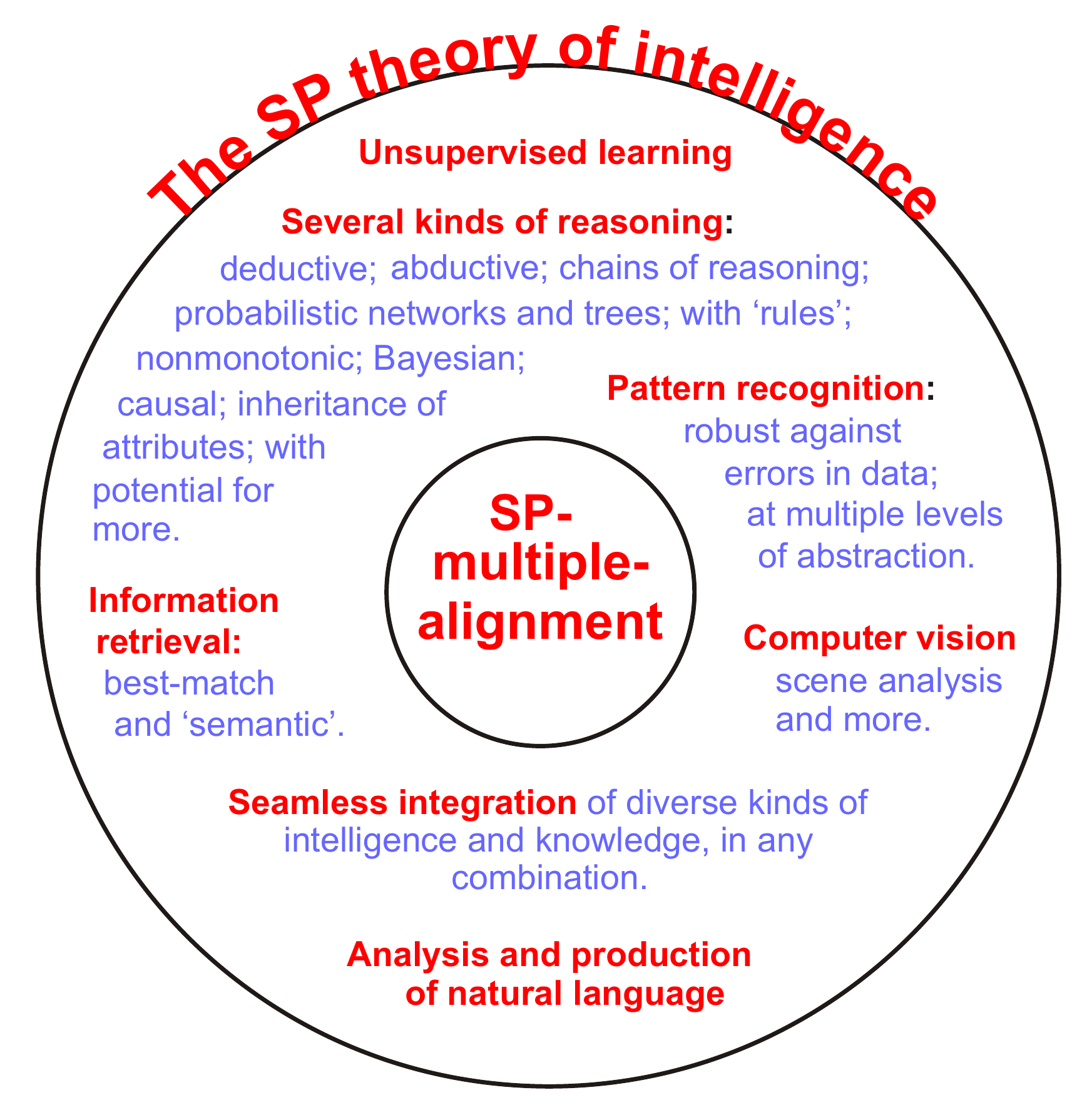}
\caption{A schematic representation of versatility and integration in the SP system, with SP-multiple-alignment centre stage.}
\label{versatility_integration_figure}
\end{figure}

The SP system, with SP-multiple-alignment centre stage, provides a much more favourable combination of conceptual simplicity with descriptive or explanatory power than any alternatives, including the many attempts to develop `unified theories of cognition' and `artificial general intelligence'. In support of that conclusion, researcher Ben Goertzel, in a review of research on the development of `artificial general intelligence', writes that ``We have not discovered any one algorithm or approach capable of yielding the emergence of [intelligence].'' \cite[p.~2]{goertzel_2012}. A schematic representation of how the SP system, with SP-multiple-alignment at centre stage, may integrate diverse aspects of intelligence and knowledge is shown in Figure \ref{versatility_integration_figure}.

Distinctive features of the SP system and its advantages compared with several AI-related alternatives are described in \cite{sp_alternatives}. In particular, Section V of that paper describes 13 problems with `deep learning' in `artificial neural networks'---the subject of much interest at present---and how, in the SP framework, those problems may be overcome. How the SP system may overcome a fourteenth problem with deep learning---`catastrophic forgetting'---is described in \cite{spdlsol_2018}.

\subsection{Organisation and Workings of the SP System}\label{organisation_and_workings_appendix}

\begin{figure}[!hbt]
\centering
\includegraphics[width=0.9\textwidth]{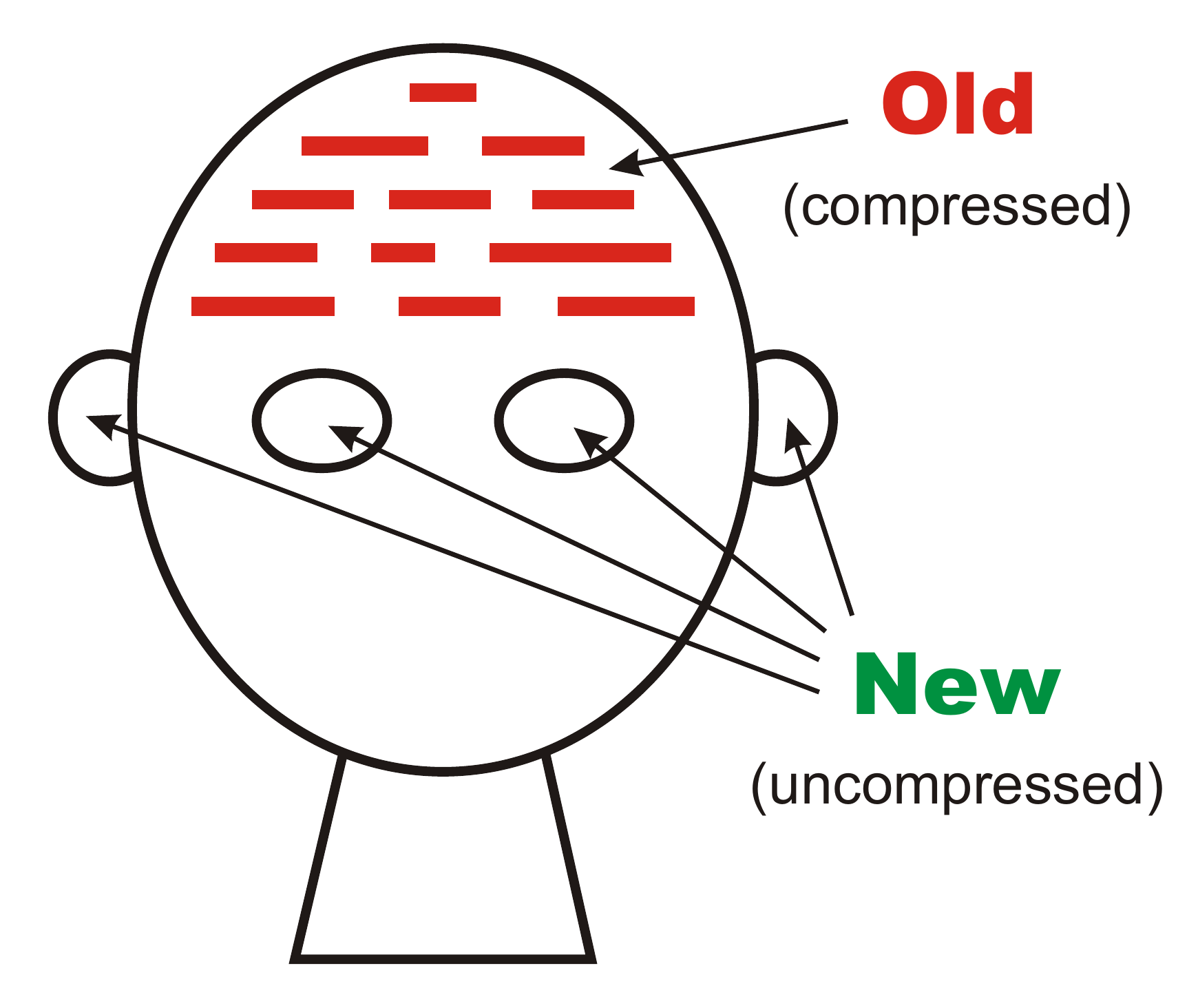}
\caption{Schematic representation of the SP system from an `input' perspective. Reproduced, with permission, from Figure 1 in \cite{sp_extended_overview}.}
\label{sp_input_perspective_figure}
\end{figure}

In broad terms, the SP system is a brain-like system that takes in {\em New} information through its senses and stores some or all of it in compressed form as {\em Old} information, as shown schematically in Figure \ref{sp_input_perspective_figure}.

All kinds of knowledge are represented in the SP system with arrays of atomic {\em SP-symbols} in one or two dimensions called {\em SP-patterns}. At present, the SP Computer Model works only with one-dimensional SP-patterns but it is envisaged that the model will be generalised to work with SP-patterns in two dimensions.

In view of evidence summarised in \cite{sp_compression}, all kinds of information processing in the SP system are done via the compression of information---via a search for patterns that match each other and the merging or `unification' of patterns (or parts of patterns) that are the same. The expression `information compression via the matching and unification of patterns' is abbreviated as `ICMUP'.

More specifically, all kinds of processing in the SP system is done via the building and processing of {\em SP-multiple-alignments} described in Appendix \ref{sp-multiple-alignments_appendix}, below. The concept of SP-multiple- alignment, as it has been developed in the SP programme of research, has the potential to be as significant for an understanding of intelligence in a broad sense as is DNA for biological sciences. It could prove to be the `double helix' of intelligence.

\subsection{SP-multiple-alignments}\label{sp-multiple-alignments_appendix}

\begin{figure}[!htbp]
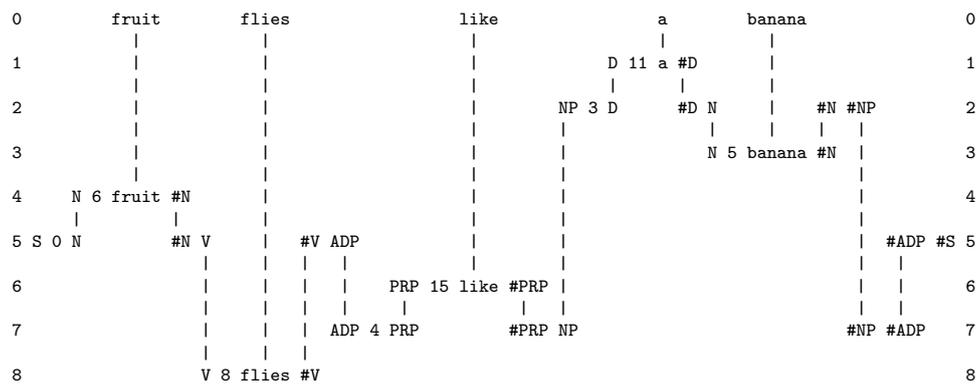

\fontsize{07.00pt}{08.40pt}
\centering
{\bf
\begin{BVerbatim}
0               fruit        flies            like              a        banana           0
                  |            |               |                |          |
1          A 12 fruit #A       |               |                |          |              1
           |          |        |               |                |          |
2     NP 2 A          #A N     |   #N #NP      |                |          |              2
      |                  |     |   |   |       |                |          |
3     |                  N 7 flies #N  |       |                |          |              3
      |                                |       |                |          |
4     |                                |       |                |    N 5 banana #N        4
      |                                |       |                |    |          |
5     |                                |       |      NP 3 D    | #D N          #N #NP    5
      |                                |       |      |    |    | |                 |
6     |                                |  V 9 like #V |    |    | |                 |     6
      |                                |  |        |  |    |    | |                 |
7 S 1 NP                              #NP V        #V NP   |    | |                #NP #S 7
                                                           |    | |
8                                                          D 11 a #D                      8

(a)

0         fruit        flies                 like                a        banana                0
            |            |                    |                  |          |
1           |            |                    |             D 11 a #D       |                   1
            |            |                    |             |      |        |
2           |            |                    |        NP 3 D      #D N     |    #N #NP         2
            |            |                    |        |              |     |    |   |
3           |            |                    |        |              N 5 banana #N  |          3
            |            |                    |        |                             |
4     N 6 fruit #N       |                    |        |                             |          4
      |         |        |                    |        |                             |
5 S 0 N         #N V     |   #V ADP           |        |                             |  #ADP #S 5
                   |     |   |   |            |        |                             |   |
6                  |     |   |   |    PRP 15 like #PRP |                             |   |      6
                   |     |   |   |     |           |   |                             |   |
7                  |     |   |  ADP 4 PRP         #PRP NP                           #NP #ADP    7
                   |     |   |
8                  V 8 flies #V                                                                 8

(b)
\end{BVerbatim}
}
\caption{The two best SP-multiple-alignments created by the SP Computer Model showing two different parsings of the ambiguous sentence {\em Fruit flies like a banana} in terms of SP-patterns representing grammatical categories, including words. Here, SP-multiple-alignments are evaluated in terms of economical encoding of information as outlined in the text. Adapted from Figure 5.1 in \protect\cite{wolff_2006}, with permission.}
\label{fruit_flies_figure}
\end{figure}

Figure \ref{fruit_flies_figure} shows two examples of SP-multiple-alignment, which may be seen as two alternative syntactic parsings of the ambiguous sentence {\em fruit flies like a banana}.\footnote{This sentence is the second part of {\em Time flies like an arrow. Fruit flies like a banana.}, attributed to Groucho Marx.}

In row 0 of each SP-multiple-alignment in Figure \ref{fruit_flies_figure}, there is a New SP-pattern representing a sentence to be parsed, with five SP-symbols, each one corresponding to a word. By convention, all New SP-patterns are shown in row 0 of each SP-multiple-alignment. Normally there is only one New SP-pattern in each SP-multiple-alignment but, as described in Section \ref{pattern_recognition_etc_section}, there can be more.

In each of rows 1 to 8 of the second of the two SP-multiple-alignments (Figure \ref{fruit_flies_figure} (b)), there is a single Old SP-pattern which represents a grammatical category such as the noun `\texttt{fruit}' in row 4, the preposition `\texttt{like}' in row 6, a noun-phrase in row 2 and the structure of a whole sentence in row 5. By convention, Old SP-patterns are always shown in rows numbered 1 and above, and there is always just one Old SP-pattern per row. The order of the Old SP-patterns across the rows is entirely arbitrary, with no special significance.

In each of these two SP-multiple-alignments, the SP-patterns represent knowledge associated with natural language. But SP-patterns, in conjunction with the SP-multiple-alignment concept, is quite general and versatile and may serve to represent and process several other kinds of knowledge, as illustrated by examples in the body of the paper.

\subsection{The Building of SP-Multiple-Alignments}\label{building_sp-multiple-alignments_appendix}

The process of building SP-multiple-alignments provides for the modelling of several different aspects of intelligence including: the analysis and production of natural language; pattern recognition that is robust in the face of errors in data; pattern recognition at multiple levels of abstraction; best-match and semantic kinds of information retrieval; several kinds of reasoning; planning; and problem solving. Learning in the SP system is somewhat different, as outlined in Appendix \ref{unsupervised_learning_appendix}.

In order to create SP-multiple-alignments like the two shown in Figure \ref{fruit_flies_figure}, the SP Computer Model must be supplied with the New SP-pattern representing the sentence to be parsed, and a set of Old SP-patterns representing a variety of grammatical structures. That set of Old SP-patterns would normally be much larger than the relatively few Old SP-patterns shown in the figure.

The overall aim is to create one or more SP-multiple-alignments where the New SP-pattern may be encoded economically in terms of the Old SP-patterns in the SP-multiple-alignment. Any such SP-multiple-alignment may be described as `good'. How the encoding is done is described in \cite[Section 4.1]{sp_extended_overview} and \cite[Section 3.5]{wolff_2006}).

Normally, the building of good SP-multiple-alignments (or good `multiple sequence alignments' in bioinformatics) is much too complicated to be achieved by any kind of exhaustive search amongst the many possibilities. Instead, it is necessary to use heuristic search, building the SP-multiple-alignments in stages, and at each stage, discarding all but the best partial structures. This approach cannot guarantee to find the best possible answer, but with enough computational resources, it can be good enough to find SP-multiple-alignments that are acceptably good.

\subsection{Unsupervised Learning}\label{unsupervised_learning_appendix}

In the SP system, learning is special. Instead of it being a by-product of the building of SP-multiple-alignments it is a process of creating `good' {\em SP-grammars}, where an SP-grammar is a collection of Old SP-patterns, and it is `good' if it is effective in the economical encoding of a target set of New SP-patterns.

That target set of New SP-patterns is the data supplied to the SP system at the start of processing, and for the sake of simplicity in this description, we shall suppose that, at the beginning of learning, the system has no Old SP-patterns. When the system has learned little or nothing, it simply adds New SP-patterns to its store of Old SP-patterns, with the addition of some `identification' symbols to each SP-pattern. But soon it can form SP-multiple-alignments via full or partial matches between New SP-patterns and Old SP-patterns. Then from SP-multiple-alignments with partial matches between SP-patterns, it may create Old SP-patterns from both matched and unmatched parts of those partially-matched SP-patterns. In this way, and via the direct addition of New SP-patterns to the stock of Old SP-patterns, the SP system may build up what may become a relatively large stock of Old SP-patterns.

In a later phase, the system tries to create one or more good grammars from its stock of Old SP-patterns. As with the building of SP-multiple-alignments, the process is normally too complex to be done by exhaustive search. As with the building of SP-multiple-alignments, the process of searching for good SP-grammars is too complex for exhaustive search so heuristic methods are needed. The system builds SP-grammars incrementally and, each stage, it discards all but the best SP-grammars.

In the SP system, learning is normally `unsupervised', deriving structures from incoming sensory information without the need for any kind of `teacher', or `reinforcement', or anything equivalent. But in case this seems unduly specialised, it appears that unsupervised learning is the most general kind of learning and that, within the framework of unsupervised learning in the SP system, there is potential to model other kinds of learning such as `supervised' learning, `reinforcement' learning, and more.


\end{document}